\documentclass[twoside,11pt]{article}

\usepackage{jmlr2e}
\usepackage[english]{babel}
\usepackage[a4paper]{geometry}
\usepackage{graphicx}
\usepackage{here}

\usepackage{hyperref}

\usepackage{amsfonts}     
\usepackage{amsmath}

\usepackage{cleveref}

\usepackage{color}
\usepackage{caption}
\usepackage{subcaption}
\usepackage{multirow}

\usepackage{./styles/mydef}

\newcommand{\edit}[1]{#1}

\ShortHeadings{On Instrumental Variable regression for Deep OPE}{Chen, Xu, Gulcehre, Le Paine, Gretton, de Freitas, Doucet}
\firstpageno{1}

\begin{document}

\title{On Instrumental Variable Regression for \\ Deep Offline Policy Evaluation}

\author{\name Yutian Chen \email yutianc@google.com\\
\addr DeepMind\\
R7, 14-18 Handyside Street\\
King’s Cross London\\
N1C 4DN
\AND
\name Liyuan Xu \email liyuan.jo.19@ucl.ac.uk\\
\addr Gatsby Unit
\AND
\name Caglar Gulcehre \email caglarg@google.com\\
\addr DeepMind
\AND
\name Tom Le Paine \email tpaine@google.com\\
\addr DeepMind
\AND
\name Arthur Gretton \email arthur.gretton@gmail.com\\
\addr Gatsby Unit
\AND
\name Nando de Freitas \email nandodefreitas@google.com\\
\addr DeepMind
\AND
\name Arnaud Doucet \email arnauddoucet@google.com\\
\addr DeepMind
}

\editor{Francis Bach}

\maketitle

\begin{abstract}

We show that the popular reinforcement learning (RL) strategy of estimating the state-action value (Q-function) by  minimizing the mean squared Bellman error leads to a regression problem with confounding, the inputs and  output  noise  being  correlated. Hence, direct minimization of the Bellman error can result in significantly biased Q-function estimates. We explain why fixing the target Q-network in Deep Q-Networks and Fitted Q Evaluation provides a way of overcoming this confounding, thus shedding new light on this popular but not well understood trick in the deep RL literature. 
An alternative approach to address confounding is to leverage techniques developed in the causality literature, notably instrumental variables (IV). We bring together here the literature on IV and RL by investigating whether IV approaches can lead to improved Q-function estimates. This paper analyzes and compares a wide range of recent IV methods in the context of offline policy evaluation (OPE), where the goal is to estimate the value of a policy using logged data only. By applying different IV techniques to OPE, we are not only able to recover previously proposed OPE methods such as model-based techniques but also to obtain competitive new techniques. We find empirically that state-of-the-art OPE methods are closely matched in performance by some IV methods such as AGMM, which were not developed for OPE. We open-source all our code and datasets at https://github.com/liyuan9988/IVOPEwithACME.
\end{abstract}


\section{Introduction}
Deep neural networks have made it possible for reinforcement learning (RL) to attain superhuman performance in challenging domains such as ATARI from raw sensory data \citep{mnih2015humanlevel} and Go \citep{silver2016mastering}. While RL is starting to be used for real-world applications \citep{bellemare2020autonomous}, its adoption remains fairly limited. Standard RL techniques require repeated interaction with the environment. This is often too costly to implement practically, and running a poor policy could lead to disastrous outcomes (e.g., in power plants or healthcare decision-making systems). While controlling a large and/or complex real-world system can be costly and risky, data acquisition is often comparatively cheap. The goal of offline RL is to evaluate and learn new policies based only on logged data, without any interaction with the environment.

In this paper, we focus on the problem of Offline Policy Evaluation (OPE), also known in the literature as Off-policy Policy Evaluation. OPE involves estimating the value of a new policy using logged data produced by possibly many different policies. This is a problem of great significance because individuals and organizations often need to choose a single policy among a wide set of proposed policies for deployment. Choosing which policy to deploy well can result in improved user satisfaction, or better medical treatments.

A plethora of methods have been proposed to address this problem; see e.g. \citet{precup2000eligibility,precup2001off,dudik2011doubly,thomas2016data,jiang2016doubly,liu2018breaking,mrdr}; see also \citet{levine2020offline,fu2021benchmarks} for recent reviews. We focus here on methods that are relying on an estimate of the state-action value function, known as the Q-function. It is well-known that the Q-function can be estimated by minimizing the mean squared Bellman error. However, the resulting regression problem is not standard as the inputs and the output noise are correlated, leading to some confounding. We show here that fixing the target Q-network in the popular Deep Q-Networks (DQN) \citep{mnih2013playing} and Fitted Q Evaluation (FQE) \citep{le2019batch} can be re-interpreted as a strategy addressing this confounding. We then investigate a different class of approaches to address the same problem. In causal inference, Instrumental Variables (IV) regression is a standard strategy for learning causal relationships between confounded treatment and outcome variables from observational data by utilizing an instrumental variable, which affects the outcome only through the treatment \citep{James2003}. The connection between RL and IV ideas was made early on by \cite{bradtke1996linear} when introducing Least Square Temporal Differences (LSTD), a method to estimate on-policy linearly parameterized value functions. Their derivation made use of the {\it two-stage least squares} (2SLS) algorithm, the most standard IV regression technique. However, the connection between RL and IV seemed to have been largely ignored ever since in the literature. 

Here we build on this connection. We exploit the fact, shown by \cite{xu2021learning}, that we can estimate a $Q$ function parameterized by a neural networks using non-linear IV regression techniques. We can thus use the non-linear IV techniques recently developed in machine learning  \citep{Hartford2017,lewis2018adversarial,Singh2019,muandet2019dual,Bennett2019,dikkala2020minimax,liao2020,xu2021learning} to perform $Q$ function estimation.

Our contributions in this paper are four-fold.
\begin{itemize}
    \item We show that estimating the state-action value ($Q$) by minimizing the mean squared Bellman error leads to a regression problem with confounding, the inputs and output noise being correlated. We provide a re-intepretation of the popular strategy consisting of fixing the target Q-network in Deep Q-Networks (DQN) \citep{mnih2013playing} and Fitted Q Evaluation (FQE) \citep{le2019batch} as a way to overcome confounding. 
    
    \item We extend the IV interpretation of the on-policy state value ($V)$ linear estimation problem to off-policy state-action value ($Q$) linear estimation. As shown recently by \cite{xu2021learning}, we can further recast the problem of non-linear $Q$-function evaluation and OPE as a non-linear IV regression problem, bringing together the literature on IV and RL.
    
    \item We review recent IV methods developed in machine learning, including Deep IV \citep{Hartford2017}, Kernel IV \citep{singh2019kernel}, Deep Generalized Method of Moments (Deep GMM) \citep{bennett2019deep}, adversarial GMM (AGMM) \citep{dikkala2020minimax} and Deep Feature IV (DFIV) \citep{xu2021learning} and specialize them to the OPE problem. By doing so, not only do we recover some OPE techniques already available, but also obtain novel methods and insights.

    \item We evaluate the performance of these techniques empirically on a variety of tasks and environments, including Behaviour Suite (BSuite) \citep{osband2019behaviour} and DeepMind Control Suite (DM Control) \citep{tassa2020dmcontrol}. We found experimentally that some of the recent IV techniques such as AGMM display performance on par with state-of-the-art FQE methods. We open-source the implementation of all methods and datasets at \url{https://github.com/liyuan9988/IVOPEwithACME}. 
    
\end{itemize}

Our main findings are that when doing OPE for a policy near to  that which generated the available data, the confounding effect can be very pronounced, and ignoring it --- as in  Deterministic Bellman Residual Minimization  (DBRM) \citep{saleh2019} --- is problematic. In this scenario, we find that the best IV method - AGMM - performs on par with FQE, and is only outperformed by distributional FQE. On more difficult scenarios where the evaluation policy is far from the behavioral policy, additional effects due to a combination of distribution shift and model mismatch come into play. In this context, while AGMM performs on par with FQE and DFQE, DBRM is also competitive, while being more stable than competing methods.

Note that there have been recent papers combining IV techniques to RL; see e.g. \cite{bennett2021off,li2021causal,liao2021instrumental}. These papers use IV methods for non-standard RL models with unobserved confounders, whereas we focus here on the standard RL model.

The rest of this paper is organized as follows. In \Cref{Sec:background}, we define the RL model of interest, and provide overviews of the offline policy evaluation problem and of instrumental variable regression. In \Cref{Sec:LSTD}, we review the LSTD method of \cite{bradtke1996linear}, introduced to estimate linearly parameterized value functions and show how it is related to 2SLS, the most popular IV method. In \Cref{Sec:nonlinearIV}, we recast the problem of non-linear $Q$ function estimation as a non-linear IV problem, and then review some of the promising recent techniques that have been developed in this context. In \Cref{Sec:experiments}, we propose two sets of benchmarking problems with stochastic environments to assess the performance of those methods in their application to OPE, compared with a state-of-the-art OPE baseline. \Cref{Sec:conclusion} concludes with a discussion.

\section{Background}\label{Sec:background}
\subsection{Reinforcement learning and offline policy evaluation}
Reinforcement learning considers a Markov decision process 
$\left\langle\mathcal{S}, \mathcal{A}, P, R, \mu_0, \gamma \right\rangle$, 
where $\mathcal{S}$ is the state space, $\mathcal{A}$ is the action space, 
and $P(s'|a,s)$ is the probability or probability density of making a transition to state $s'$ when taking action $a$ in state $s$.
$R(r|s,a)$ denotes the probability density of observing reward $r$ after having taken action $a$ from $s$,
and $\mu_0(s)$ is the initial state distribution.
Let $\pi$ be a policy of an agent, and denote $\pi(a|s)$ as the probability or probability density of selecting action $a$ in state $s\in\mathcal{S}$.
With a discount factor $\gamma \in (0,1]$, 
the state-action value function - i.e. $Q$ function - is defined by
\begin{equation}
    Q(s,a)=\expect{\sum_{t=0}^\infty \gamma^t r_t | s_0=s, a_0=a},
\end{equation}
with $a_t\sim \pi(\cdot \mid s_t), s_{t+1}\sim P(\cdot | s_t, a_t), r_t \sim R(\cdot | s_t, a_t)$ for $t\geq 0$. 

Common tasks in reinforcement learning including estimating the \emph{value} of a given target policy $\pi$ or optimizing the policy value with respect to $\pi$, where the policy value is defined by the expected sum of discounted rewards from the initial state distribution
\begin{equation}
\rho(\pi)=\expect[s_0\sim \mu_0]{\sum_{t=0}^\infty \gamma^t r_t}=\mathbb{E}_{s\sim \mu_0, a|s\sim \pi}[Q(s,a)],
\label{eq:policy_value}
\end{equation}

When an agent is prohibited to interact with the environment directly, one has to rely on an existing dataset of trajectories or transition tuples $(s, a, r, s')$, to estimate the policy value or learn the optimal policy. The dataset could have been collected by one or a mixture of potentially unknown policies of potentially unknown analytical form, denoted by $\pi_b(\cdot|s)$, and the corresponding state action distribution is denoted by $\mu_b$.

The goal of offline policy evaluation (OPE) is to evaluate the value of a target policy, $\pi$, based on the offline behavior dataset.
This problem has been extensively studied in the literature. The readers are referred to \citet{levine2020offline} for a review and \citet{voloshin2019empirical,fu2021benchmarks} for benchmarks of recent OPE algorithms.

One family of OPE approaches is to estimate the value function based on the Bellman equation,
\begin{equation}\label{eq:BellmanQ}
    Q(s,a)=\expect[r\sim R(\cdot|s, a)]{r|s,a}+\gamma \expect[s'\sim P(\cdot|s, a), a'\sim \pi(\cdot|s')]{Q(s',a')|s, a}, \forall s\in \mathcal{S}, a\in\mathcal{A}.
\end{equation}
We can solve the Bellman equation as a least square regression problem for the reward $r$ on the state-action pair $(s, a)$
\begin{align}
	\label{eq:regress}
    \expect{r|s, a} = Q(s,a) - \gamma \expect{Q(s',a')|s, a},
\end{align}
and find the function $Q$ to minimize the mean squared Bellman error (MSBE) \citep[p. 268]{Sutton2018} with respect to the behavior distribution
\begin{align}\label{eq:def-msbe}
    Q = \underset{Q}{\arg\min}~\expect[(s, a)\sim \mu_b, r\sim R]{\left(r - Q(s, a) + \gamma \expect[s'\sim P(\cdot|s,a), a'\sim \pi(\cdot | s')]{Q(s', a')}\right)^2}. 
\end{align}
We then use \Cref{eq:policy_value} for the evaluation policy $\pi$.

\subsubsection{A Simple Biased Estimator}
A simple algorithm to approximate the objective of \Cref{eq:def-msbe} using transition samples $(s, a, r, s', a')$ from the dataset is known as Deterministic Bellman Residual Minimization (DBRM) \citep{saleh2019}. We consider here a simple variant of DBRM as a baseline with two independent action samples from the given target policy,
\begin{align}\label{eq:dbrm}
    &Q_{\text{DBRM}} = \underset{Q}{\arg\min} ~\expect{\left(r - Q(s, a) + \gamma {Q(s', {a'}^{(1)})}\right)\left(r - Q(s, a) + \gamma {Q(s', {a'}^{(2)})}\right)}\,,
\end{align}
where $(s, a)\sim \mu_b, r\sim R, s'\sim P(\cdot|s,a), {a'}^{(1)}\sim \pi(\cdot | s'), {a'}^{(2)}\sim \pi(\cdot | s')$.
The argument within the expectation of \Cref{eq:dbrm} is an unbiased estimate of MSBE only if the MDP's transition dynamics $P$ and the target policy $\pi$ are deterministic.  If this is not the case, we would require two independent samples of $s'$ starting from the same $(s,a)$ to obtain an unbiased estimate of the MSBE objective in \Cref{eq:def-msbe} \citep{baird1995residual}. This is usually not possible. More practical and sophisticated methods have been proposed to mitigate the bias \citep{antos2008learning,munos2008finite}.

\subsubsection{Fitted Q Evaluation}\label{sec:fqe}
Alternatively, one can move the troublesome expectation in \Cref{eq:regress} from inside the regression function to the target as follows,
\begin{align}
	\label{eq:regress_fqe}
    \expect{r|s, a} = Q(s,a) - \gamma \expect{Q(s',a')|s, a} \Longrightarrow \expect{r + \gamma Q(s',a')|s, a} = Q(s,a),
\end{align}
and minimize the least squared temporal difference (TD) error iteratively, with the $Q$ function on the left hand side being fixed at every iteration,
\begin{equation}\label{eq:fqe}
	Q_k = \underset{Q}{\arg\min} 
		~\expect[(s, a)\sim \mu_b, s'\sim P, a'\sim \pi, r\sim R]{\left(Q(s, a) - \left(r + \gamma {Q_{k-1}(s', a')}\right) \right)^2}.
\end{equation}
This method is known as fitted $Q$ evaluation (FQE) \citep{le2019batch}, a variant of the fitted $Q$ iteration \citep{ernst2005tree} algorithm.
\citet{fu2021benchmarks} show that FQE outperforms other OPE algorithms in a deep OPE benchmark.
The same idea was used in other approximate dynamic programming approaches such as the Deep Q Network (DQN) \citep{mnih2015humanlevel} where the parameters of the target Q network (corresponding to $Q_{k-1}$ in \Cref{eq:fqe}) were fixed when updating the online Q network.

\subsection{Instrumental variable regression}
{\it Instrumental variable} (IV) regression methods \citep{James2003} are standard techniques developed in the causal inference and econometrics literature, which are used to predict the effect of actions $X$ (called \textit{treatment}) on the world when the treatment affects the distribution of the variable of interest $Y$, which is called the \textit{outcome}. IV regression provides a framework to assess this effect (called the \textit{structural function}) by using an instrumental variable $Z$, which only affects the treatment directly, but has no direct effect on the outcome. 

Instrumental variables can be found in many contexts, and IV regression is extensively used by economists and epidemiologists. For example, \citep{Wright1928, Blundell2012} used supply cost shifters as instrumental variables, and estimate the effect of price on the demand to correct for confounders such as the time of the year. IV regression has also been used to measure the effect of a drug in the scenario of imperfect compliance \citep{Angeris1996}, or the influence of military service on lifetime earnings \citep{Angeris1990}.

Formally, we aim  to learn a relationship between $X$ and $Y$, which is generated from 
\begin{align}\label{eq:structuralcausalmodel}
    Y = f_\mathrm{struct}(X) + \varepsilon, \quad \expect{\varepsilon} = 0, \quad \expect{\varepsilon | X} \neq 0,
\end{align}
where $f_\mathrm{struct}$ is the structural function, which we assume to be continuous, and $\varepsilon$ is an additive noise term. The challenge is that $\expect{\varepsilon | X} \neq 0$, which reflects the existence of a latent confounder. Hence, we cannot use ordinary supervised learning techniques since $f_\mathrm{struct}(x) \neq \expect{Y | X=x}$. 

\begin{figure}[tbhp!]
    \centering
    \includegraphics[width=0.5\textwidth]{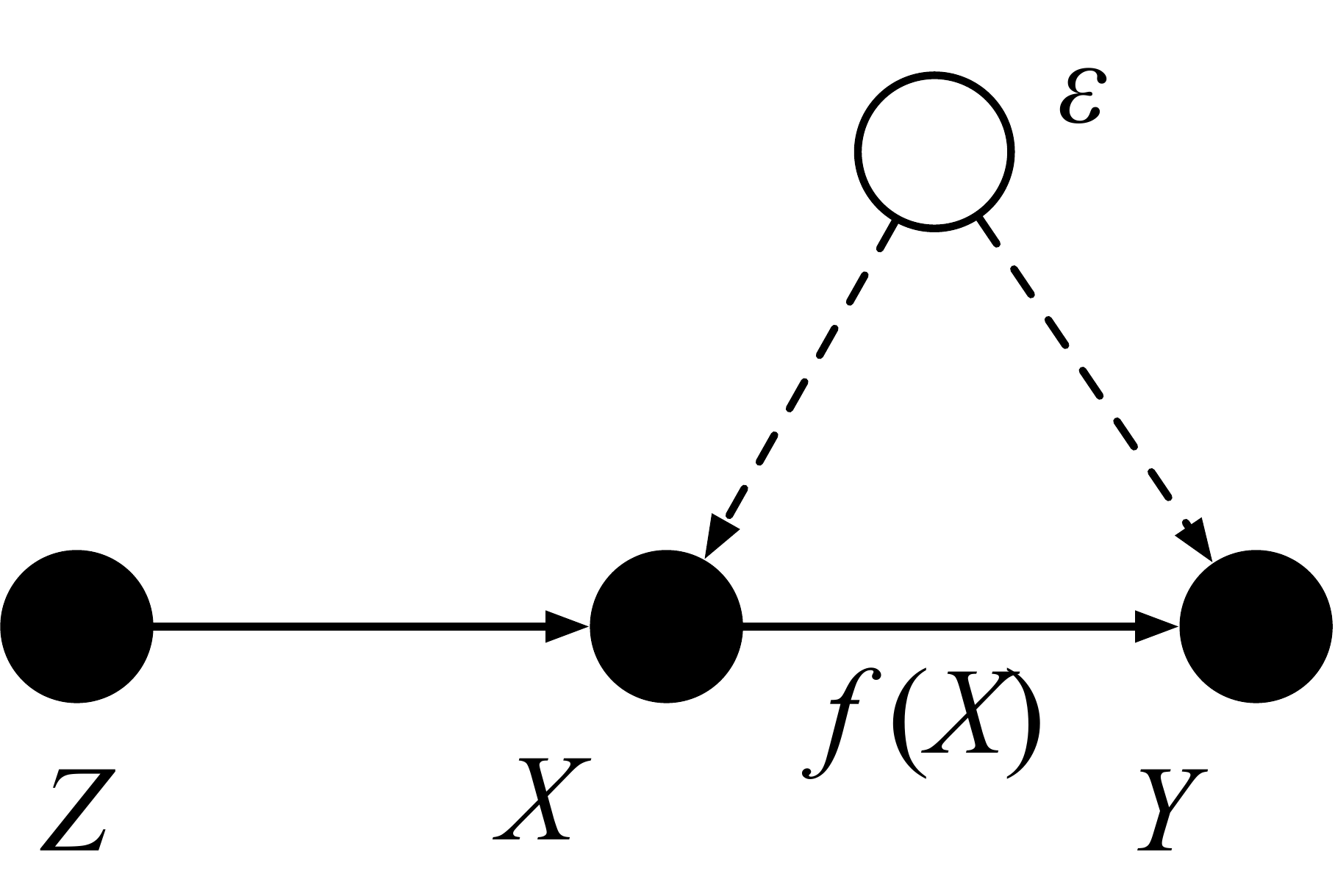}
    \caption{Causal graphical model for instrumental variable methods.}
    \label{fig:iv}
\end{figure}

To deal with the confounder $\varepsilon$, we assume to have access to an instrumental variable $Z \in \mathcal{Z}$ which satisfies the following assumption. 
\begin{assum}\label{assum:iv}
    The conditional distribution $P(X | Z)$ is not constant in $Z$ and one has $\expect{\varepsilon | Z} = 0$.
\end{assum}
Intuitively, Assumption~\ref{assum:iv} means that the instrument $Z$ induces variation in the treatment $X$ but is uncorrelated with the hidden confounder $\varepsilon$. The causal graph describing these relationships is shown in Figure~\ref{fig:iv}.\footnote{We show the simplest causal graph in Figure~\ref{fig:iv}
It entails $Z \indepe \varepsilon$,
but we only require $Z$ and $\varepsilon$ to be uncorrelated in Assumption~\ref{assum:iv}. Of course, this graph also says that $Z$ is not independent of $\varepsilon$ when conditioned on  observations $X$.} Note that the instrument $Z$ cannot have an incoming edge from the latent confounder that is also a parent of the outcome.  It follows directly from \Cref{assum:iv} that
\begin{align}\label{eq:deep_iv}
    \expect{Y|Z} = \expect{f(X)|Z} + \expect{\varepsilon|Z}
    = \int_X f(X) P(X|Z) \text{d}X.
\end{align}

Classically, IV regression is solved by the {\it two-stage least squares} (2SLS) algorithm; we learn a mapping from the instrument to the treatment in the first stage, and learn the structural function in the second stage as the mapping from the conditional expectation of the treatment given the instrument (obtained from stage 1) to the outcome. Originally, 2SLS assumes linear relationships in both stages, i.e.,
$$X = Z \omega + \delta,\quad f_\mathrm{struct}(X) = X \theta\,,$$
where $\omega$ and $\theta$ are unknown regression coefficients and $\delta$ is a random variable satisfying $\expect{\delta|Z} = 0$.
In the case when the dimension of $X$ equals that of $Z$, the 2SLS estimator has the following simple form
\begin{equation}\label{eq:2sls}
    \hat{\theta} = \left(Z^\top X\right)^{-1}\left(Z^\top Y\right)\,,
\end{equation}
where we somewhat abuse notation and denote by $X$, $Z$ the matrices of observed treatment and instrumental variables, by $Y$ the vector of outcomes, and each row corresponds to one observation.

Instrumental variable methods have been extended to non-linear settings first in economics \citep{Newey2003,Carrasco2007,Darolles2011,Blundell2012,Chen2018,} and more recently in machine learning  \citep{Hartford2017,lewis2018adversarial,Singh2019,muandet2019dual,Bennett2019,dikkala2020minimax,liao2020,xu2021learning}. One approach has been to use non-linear feature maps.  Sieve IV \citep{Newey2003, Chen2018} uses a finite number of explicitly specified basis functions. Kernel IV (KIV) \citep{Singh2019} and Dual IV regression \citep{muandet2019dual} extend sieve IV to allow for an infinite number of basis functions using reproducing kernel Hibert spaces (RKHS). Although these methods enjoy desirable theoretical properties, the flexibility of the model is limited due to the prespecified features. To mitigate this limitation, \citet{xu2021learning} proposes Deep Feature IV (DFIV) method, in which one learns features adaptively using neural networks while preserving the two-stage nature of 2SLS.

Another non-linear approach to the stage 1 regression is to estimate the conditional distribution $P(X|Z)$ \citep{Carrasco2007, Darolles2011,Hartford2017}. This allows flexible models, including deep neural nets, as proposed in the DeepIV algorithm of \citep{Hartford2017}. However, the conditional density estimation can be costly, and can suffer from high variance when the treatment is high-dimensional.

As a further alternative, several recent works \citep{lewis2018adversarial,Bennett2019,dikkala2020minimax,liao2020} have been inspired by another instrumental variable technique, the Generalized Method of Moments (GMM) \citep{Hansen1982}, and find non-linear structural functions to ensure that the regression residual and the instrument are uncorrelated. These works do not require two stage regression, and the resulting methods are often formulated as solving a minimax optimization problem.

\section{Relationship between LSTD and Linear IV}\label{Sec:LSTD}
Solving \Cref{eq:def-msbe} requires computing the conditional expectation $\expect{Q(s', a')|s,a}$. This is usually infeasible in practice because it would require being able to reset the environment to state $s$ and draw multiple samples of the next state $s'$.
\citet{bradtke1996linear} proposed the Least Square Temporal Difference (LSTD) algorithm to solve this problem with a single sample of $s'$. Specifically, they consider estimating the \textit{state value function} $V(s) = \expect[a\sim \pi]{Q(s, a)}$ in the \textit{on-policy} RL setting, i.e., $\pi_b = \pi$, and assume the value function can be parameterized as a linear function of a fixed set of features. \citet{bradtke1996linear} pointed out originally that \Cref{eq:def-msbe} could be reformulated and solved with a linear IV method.

\citet{lagoudakis2003least} proposed to model the \textit{state-action value function} $Q$ instead of $V$ using a similar linear combination of features, and extended LSTD to LSTD-Q so that it could be applied to off-policy RL, $\pi_b \neq \pi$. Their derivation does not rely on IV ideas. 
We propose here an alternative derivation of LSTD-Q which is a natural extension of the IV approach pioneered in \citet{bradtke1996linear} to the $Q$ function.

\begin{figure}[tbhp!]
    \centering
    \includegraphics[width=0.5\textwidth]{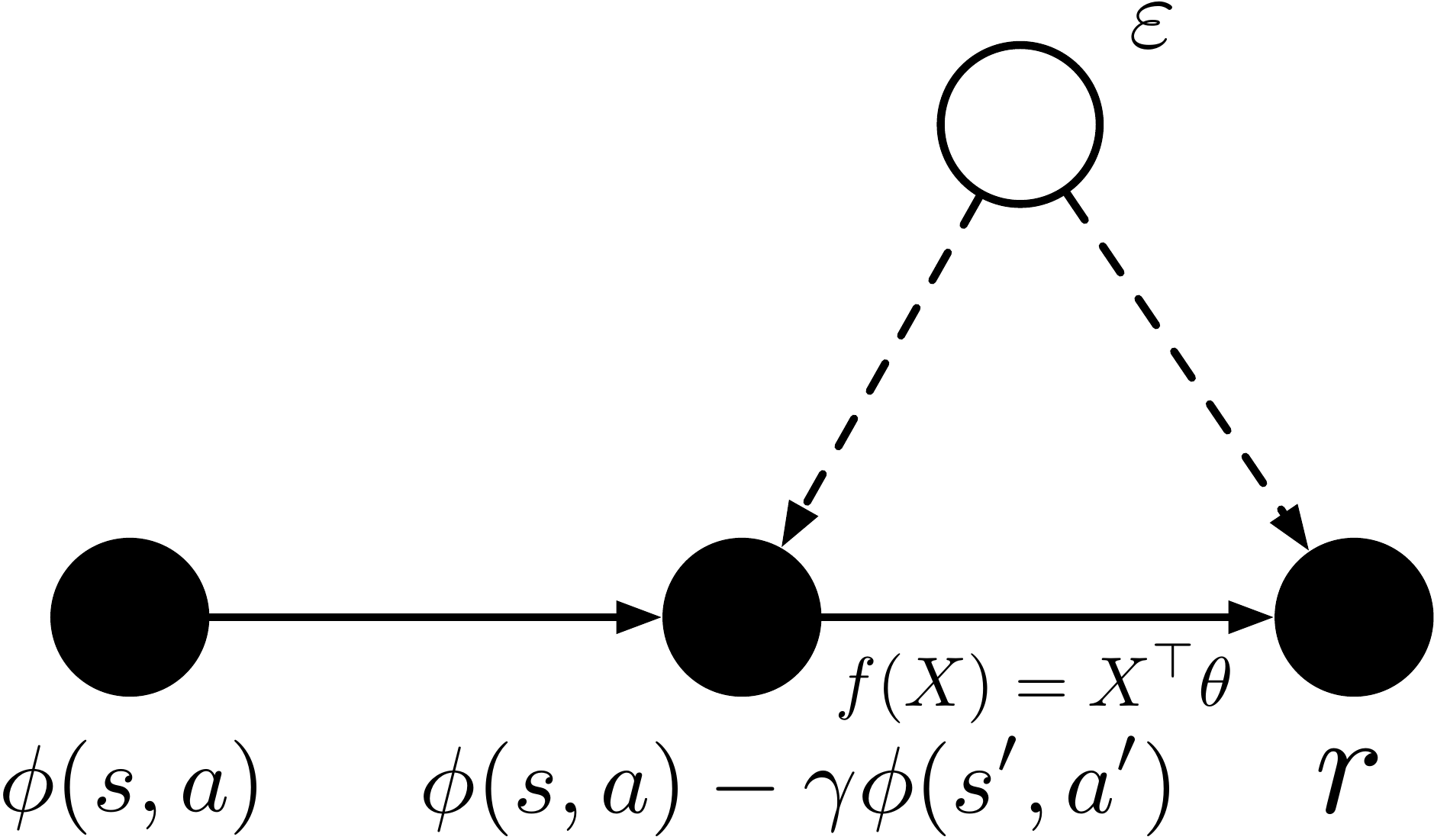}
    \caption{Causal graphical model of LSTD}
    \label{fig:lstd}
\end{figure}

We consider a linear approximation of the form 
\begin{equation}\label{eq:linearapprox}
    Q(s,a)=\phi(s,a)^\top~\theta,
\end{equation}
where $\phi(s,a)$ is a set of features evaluated at $(s,a)$ and $\theta$ is the parameter vector to estimate.
\emph{If \Cref{eq:linearapprox} were exact and not an approximation}, then we could rewrite \Cref{eq:regress} as 
\begin{align}
 \underbrace{r}_{Y} &= Q(s,a) - \gamma Q(s',a') + \left(\gamma Q(s',a') - \gamma \expect{Q(s',a')|s, a}\right) + \left(r - \expect{r|s, a}\right) \label{eq:iv_regression} \\
   &={\underbrace{\left(\phi(s,a)-\gamma \phi(s',a')\right)}_{X}}^\top~\theta
     + \underbrace{\gamma\left(\phi(s',a') - \expect{\phi(s', a')|s, a}\right)^\top~\theta
     + r-\expect{r|s,a}}_{\varepsilon} , \label{eq:lstd_regression}
\end{align}
where $(s, a) \sim \mu_b,~~ s' \sim P(s'|s, a),~~ a'\sim \pi(a'|s')$. The decomposition in \Cref{eq:iv_regression} was discussed in \cite{xu2021learning}. \Cref{eq:lstd_regression} matches the regression formulation in \Cref{eq:structuralcausalmodel} (see \Cref{fig:lstd} for the causal graphical model).

Because the observation noise $\varepsilon$ and input $X$ are correlated through the sample of $s', a'$, we have a confounded regression problem, that is
\begin{align}
& \expect{\varepsilon} = 0, \text{ but } \nonumber\\
& \expect{\varepsilon|X} \neq 0, \text{ because } \expect{\phi(s',a') - \expect{\phi(s',a')|s,a}|s, a, s', a'} \neq 0 \,.
\end{align}
Solving the least squared minimization problem of \Cref{eq:lstd_regression} directly will lead to an inconsistent estimate of $\theta$.
By choosing the feature of $(s, a)$ as the instrumental variable, $Z = \phi(s, a)$, we can show that $Z$ is uncorrelated with both terms in $\varepsilon$
\begin{align}
\text{corr}(\phi(s, a), \phi(s',a') - \expect{\phi(s',a')|s,a}) = 0, \quad \text{corr}(\phi(s, a), r-\expect{r|s,a}) = 0\,,
\label{eq:lstd_uncorrelation}
\end{align}
which follows directly from Lemma 4 in \cite{bradtke1996linear} by swapping $s$ for $(s, a)$. Therefore, we have $\expect{\varepsilon|Z} = 0$.
In this scenario \Cref{assum:iv} for applying IV regression is satisfied. We can thus derive the same LSTD-Q estimator using 2SLS as
\begin{equation}
\hat{\theta}  = \left(\Phi^\top \left(\Phi - \gamma \Phi'\right)\right)^{-1} \Phi^\top R \,,
\end{equation}
where $\Phi, \Phi'$, and $R$ are matrices where every row corresponds respectively to the transpose of $\phi(s, a), \phi(s', a')$ and $r$ from the offline dataset, except for $a' \sim \pi(a'|s')$.

We note that the derivation by \cite{bradtke1996linear} requires that the value function lives in the linear subspace of the features, \Cref{eq:linearapprox}. Then they show that as the number of data increases, the solution converges under regularity conditions to the least-square fixed-point approximation (without mentioning it). The instrumental variable interpretation also requires the structural function in \Cref{eq:lstd_regression} to hold, which is derived from \Cref{eq:linearapprox}. Convergence of the LSTD algorithm does not require \Cref{eq:linearapprox} to be valid. \cite{lagoudakis2003least} show that there is indeed no need to make such an assumption, and obtain a direct derivation by least-square fixed-point approximation. We refer the readers to \cite{lagoudakis2003least} for this alternative interpretation.

\begin{remark}
In the formulation of FQE in \Cref{sec:fqe}, because $Q(s', a')$ is part of the output, the resulting regression problem becomes
\begin{align}\label{eq:fqe_linear}
 \underbrace{r + \gamma \phi(s',a')^\top~\theta}_{Y}
   &={\underbrace{\phi(s,a)}_{X}}^\top~\theta
     + \underbrace{\gamma\left(\phi(s',a') - \expect{\phi(s', a')|s, a}\right)^\top~\theta
     + r-\expect{r|s,a}}_{\varepsilon} \,.
\end{align}
One can see that this regression problem is not confounded as $\varepsilon$ is uncorrelated with $X$ (see \Cref{eq:lstd_uncorrelation}), but $\theta$ in the target $Y$ has to be fixed when estimating the parameter.
\end{remark}

\section{Policy Evaluation with Non-linear Functions and Non-linear IV}\label{Sec:nonlinearIV}
In this section, we will first extend the LSTD algorithm to the scenario with a non-linear value function and formulate it as a non-linear IV problem. We then introduce a few recent representative non-linear IV methods as OPE algorithms under that setting, using our notations for consistency whenever possible.

\subsection{Extension to Non-linear Value Functions}
\begin{figure}[tbhp!]
    \centering
    \includegraphics[width=0.5\textwidth]{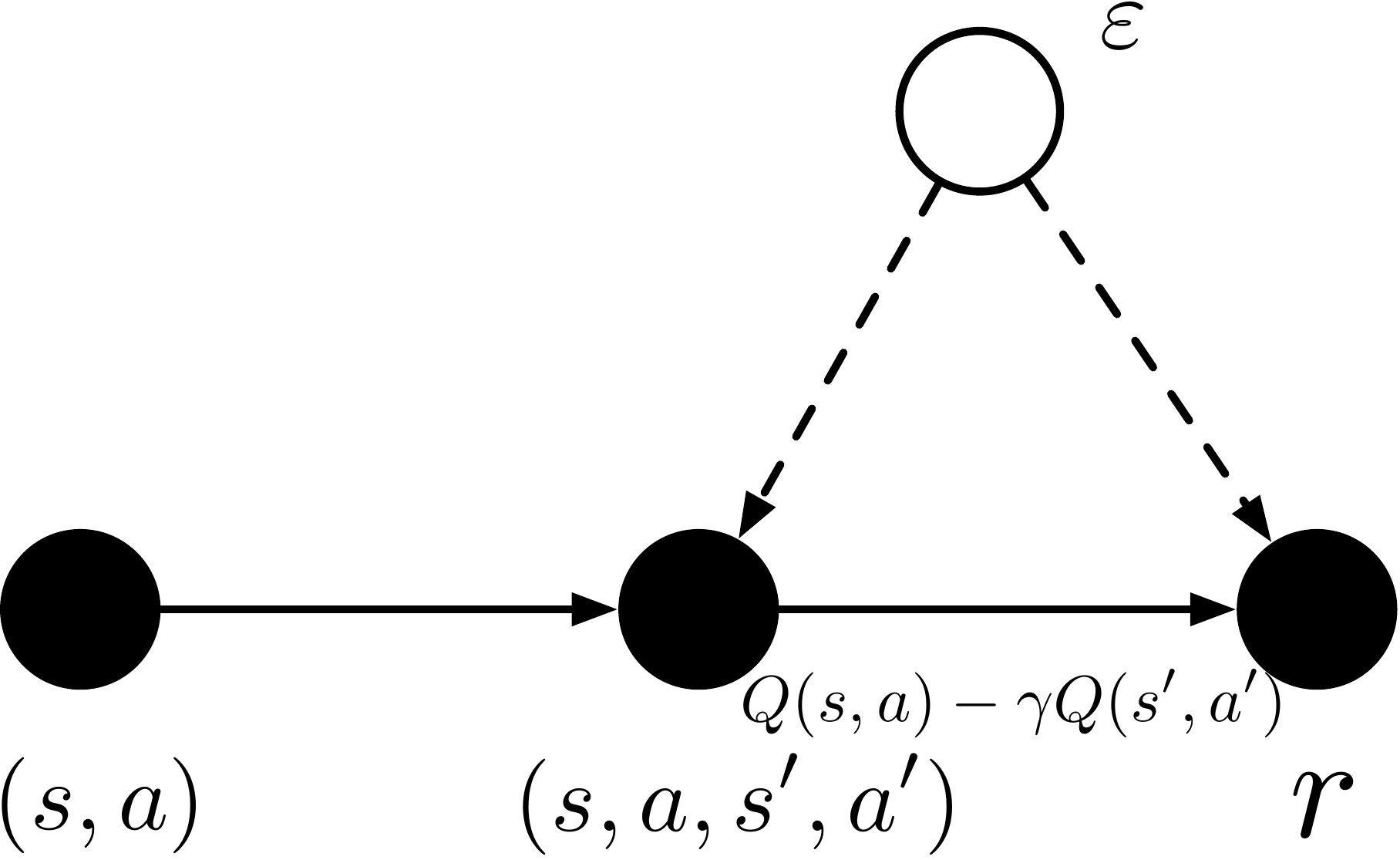}
    \caption{Causal graphical model for policy evaluation with non-linear $Q$ function.}
    \label{fig:nonlinear_iv}
\end{figure}

When the value function $Q$ is a non-linear function of $(s, a)$, it was shown recently by \cite{xu2021learning} that we can estimate it by solving a non-linear IV regression problem following \Cref{eq:iv_regression} with the corresponding causal graphical model in \Cref{fig:nonlinear_iv},
\begin{align}
 \underbrace{r}_{Y} &= \underbrace{Q(s,a) - \gamma Q(s',a')}_{f(X)} + \underbrace{\left(\gamma Q(s',a') - \gamma \expect{Q(s',a')|s, a}\right) + \left(r - \expect{r|s, a}\right)}_{\varepsilon}, \label{eq:nonlinear_regression}
\end{align}
where $(s, a) \sim \mu_b,~~ s' \sim P(s'|s, a),~~ a'\sim \pi(a'|s')$ and the structural function is
$f(s, a, s', a') = Q(s,a) - \gamma Q(s',a')$ with $X=(s, a, s', a')$. Choosing the instrument $Z=(s,a)$, it is easy to show that $Z$ satisfies \Cref{assum:iv} as $P(X|Z)$ is not constant in $Z$, where $Z$ is a subset of $X$ and $\mathbb{E}(\varepsilon|Z)=0$.

In practice, we need to parameterize the structural function $f(s,a,s',a')$. It is sensible to use a parameterization of the form 
\begin{equation}
f_{\theta}(s,a,s',a')=Q_{\theta}(s,a)-\gamma Q_{\theta}(s',a')
\end{equation}
instead of parameterizing $f$ directly. We should also keep in mind that when we use approximate $Q$ functions it will not be true that $\mathbb{E}(\varepsilon|Z)=0$ exactly. The induced bias will be illustrated later in the experiments with an under-fitted function approximator.

The dependency of the output noise $\varepsilon$ on the next state $s'$ bears resemblance to the ``colored'' noise in the GPTD formulation \citep{engel2005reinforcement}. However, we emphasize here that the GPTD formulation fails to reflect the confounding issue inherent to the Bellman residual minimization problem because the noise $N(s, s')$ in GPTD is still assumed to be uncorrelated with $s'$ even though dependent. Therefore, their solution may still lead to a biased estimator.

Depending on the function family of $Q_\theta$ and how to apply the instrumental variables, we will present a few representative non-linear IV methods in the setting of offline policy evaluation in the following sections.

\subsection{Deep IV}
The Deep IV method in \cite{hartford2017deep} is based on the identity \Cref{eq:deep_iv}
\begin{align}\label{eq:deep_iv2}
    \expect{Y|Z} = \int_X f(X) P(X|Z) \text{d}X.
\end{align}
\edit{It is a two-stage regression approach that estimates the nonlinear relationships between $Z$ and $X$, and $X$ and $Y$, using a neural network function approximator  in each stage.} In the first stage, Deep IV trains a treatment network to estimate the conditional distribution of treatment $P_{\phi}(X|Z)$ by maximum likelihood estimation. The network outputs a categorical distribution if the treatment variable $X$ is discrete and a mixture of Gaussian distributions if it is continuous. In the second stage, it estimates the structural function using an outcome network $f_\theta(X)$ by regressing $Y$ on the conditional expectation $\expect[P_\phi(X|Z)]{f_\theta(X)|Z}$ where the estimate of the expectation is obtained by Monte Carlo samples from $P_{\phi}(X|Z)$.

In our RL context, recall that $X=(s,a,s',a')$ while $Z=(s,a)$, so the non-degenerate part of the conditional distribution of the treatment is given by
\begin{equation}
    P(s',a'|s,a)=P(s'|s,a)\pi(a'|s').
\end{equation}
Thus, applying Deep IV in the RL context consists first of estimating the transition distribution $P(s'|s,a)$ using a generative model (as we know the target policy $\pi(a'|s')$). We then compute a Monte Carlo estimate of $\mathbb{E}(f_{\theta}(X)|Z)$ with multiple samples, and estimate $Q$ by minimizing the approximate MSBE in \Cref{eq:def-msbe}.

\cite{hartford2017deep} also allow for observable confounders, but this extension is unnecessary in our scenario. Deep IV is closely related to the model-based reinforcement learning algorithm Dyna-Q \citep{sutton1990integrated}, which also learns the transition distribution and then $Q$ by minimizing the TD error. The difference is that Dyna-Q uses Q-learning to minimize the TD error in the Bellman optimal equation in order to improve the policy instead of estimation, and it requires only a single sample of $s'$ to provide an unbiased estimate of the gradient, similar to the iterative FQE algorithm.

\subsection{KIV and DFIV}
KIV \citep{Singh2019} and DFIV \citep{xu2021learning} \edit{introduce nonlinear feature maps in the IV formulation. Similar to Deep IV, this approach is also based on \Cref{eq:deep_iv}, and solves for $f$ by minimizing}
\begin{align*}
    \mathcal{L}(f) = \expect[YZ]{(Y - \expect[X|Z]{f(X)})^2} + R(f), \label{eq:total-loss}
\end{align*}
\edit{where $R(f)$ is the regularization for $f$. KIV and DFIV regress to the expected features of $X$ on $Z$, however, in contrast to Deep IV, which estimates the conditional distribution of $P(Z|X)$; density estimation is not necessary for estimating $\expect[X|Z]{f(X)},$ and can be more difficult in practice. KIV and DFIV employ the following models:
\begin{align*}
    f(X) = w^\top \phi(X), \quad \expect{\phi(X)|Z} = V \psi(Z),
\end{align*}
where $w, V$ are the learnable linear weights and $\phi(X), \psi(Z)$ are the nonlinear feature maps for the treatment and the instrument, respectively.}

\edit{KIV considers  static feature maps from a Reproducing Kernel Hilbert Space (RKHS) for $\phi(X), \psi(Z)$ and learns weights $w, V$  by a two-stage regression: stage 1 performs the regression from the instrument \edit{$Z$} to the treatment features \edit{$\phi(X)$ to learn weight $V$}; then in stage 2,  weights $w$ are learned by minimizing the loss $\mathcal{L}(f)$ using the predicted treatment features $V \psi(Z)$. DFIV additionally learns feature maps $\phi(X), \psi(Z)$ using neural networks in the same two-stage regression.}

\edit{Note that in the RL context where $X=(s,a,s',a')$ and $Z = (s,a)$, the loss $\mathcal{L}$ is identical to MSBE defined in \cref{eq:def-msbe} apart from the regularization term.} This two-stage regression proceeds as follows. As in \cref{eq:linearapprox}, KIV and DFIV model $Q(s,a) = \phi(s,a)^\top \theta$ where $\phi$ is a feature map and $\theta$ are the parameters. Furthermore, they model the conditional expectation as $\expect[s',a'|s,a]{\phi(s',a')} = V\psi(s,a)$, where $\psi(s,a)$ is another feature map and $V$ is the parameter matrix to be learned. 

In stage 1, $V$ is learned by minimizing the following loss,
\begin{align}
    \hat{V} = \argmin_{V} \mathcal{L}_{1}(V), \quad \mathcal{L}_{1}(V) = \expect[s,a,s',a']{\|\phi(s',a') - V\psi(s,a) \|^2} + \lambda_1 \|V\|^2, 
\end{align}
where $\lambda_1>0$ is a regularization parameter. This is a linear ridge regression problem with multiple targets, which can be solved analytically. In stage 2, given $\hat{V}$, $\theta$ is obtained by minimizing the loss
\begin{align}
\hat{\theta} = \argmin_\theta \mathcal{L}_{2}(\theta), \quad \mathcal{L}_{2}(\theta) = \expect[r,s,a]{\|r - \theta^\top (\phi(s,a) - \hat{V}\psi(s,a)) \|^2} + \lambda_2 \|\theta\|^2, 
\label{eq:kiv_stage2_loss}
\end{align}
where $\lambda_2>0$ is another regularization parameter. Stage 2 corresponds to a ridge linear regression from $\phi(s,a) - \hat{V}\psi(s,a)$ to $r$, and also has a closed-form solution. 

In KIV, from the characteristics of RKHS functions one can learn a non-linear $Q$ function while retaining the closed-form solution of the two-stage regressions. In DFIV, because of the use of adaptive features for $\phi(s,a)$ and $\psi(s,a)$ parameterized with neural networks they learn those features by alternating the two regression stages, which enables one to learn a more flexible $Q$ function compared to KIV. In this paper, we consider a variant of DFIV that regresses $\phi(s,a) - \gamma \phi(s',a')$ instead of $\phi(s',a')$ on instrumental variables in stage 1, with details explained in \Cref{Sec:dfiv_variant}. We found this approach is more stable than the original version in the experiments. While we can derive a similar variant for KIV, we did not notice an improvement in performance.

\subsection{Generalized Method of Moments}
A family of non-linear IV methods is based on the moment restrictions derived from \Cref{assum:iv} of instrumental variables,
\begin{equation}
    \label{eq:moment_restriction}
    \mathbb{E}(\varepsilon|Z)=\mathbb{E}(Y - f(X)|Z)=0, \forall Z.
\end{equation}
\edit{In the linear setting, we require the following unconditional moment to be zero, whose solution is \Cref{eq:2sls},
\begin{equation}
    \mathbb{E}(\varepsilon Z)=\mathbb{E}[(Y - f(X))Z]=0.
\end{equation}
In the non-linear setting,} by defining a set of potentially infinitely many test functions of $Z$, $g \in \mathcal{G}$, we require all the unconditional moments to be zero,
\begin{equation}
    \Psi(f, g) = \expect[X,Y,Z]{(Y - f(X))g(Z)} = 0, \forall g.
\end{equation}

The unified solution in the family of Generalized Method of Moments (GMM) \citep{Hansen1982} is a saddle-point of the following minimax objective function,
\begin{equation}
    \label{eq:gmm_objective}
    f^* = 
    \underset{f\in\mathcal{F}}{\arg\inf}
    \sup_{g\in\mathcal{G}} \Psi_n(f, g) + R_f(f) - R_g(g)\,,
\end{equation}
where \edit{$g$ is optimized to find the largest violation of the moment condition for the current estimate of $f$} and the expectation in $\Psi$ is the empirical estimate from a dataset of size $n$. $R_f(f)$ and $R_g(g)$ are regularization terms for $f$ and $g$, respectively, for identifiability.

In the context of RL, this objective is thus given by
\begin{align}
    Q^* &= 
    \underset{Q\in\mathcal{Q}}{\arg\inf}
    \sup_{g\in\mathcal{G}} \Psi_n(Q, g) + R_f(Q) - R_g(g)\,, \nonumber \\
    \text{with } \Psi_n(Q, g) &= \expect[s,a\sim \mu_b, s'\sim P, r\sim R, a'\sim\pi]{(r - Q(s, a) + \gamma Q(s', a'))g(s, a)}.
    \label{eq:gmm_objective_q}
\end{align}
Here $g$ acts to find the largest moment of the TD error. Variants of GMM method differ in the choice of the function space~$\mathcal{Q}$, $\mathcal{G}$ and the regularization functions.

\subsubsection{Deep GMM}
Inspired by the optimally weighted GMM method in linear IV problems, \cite{bennett2019deep} propose the Deep GMM method. 
\edit{In the linear setting with a fixed set of feature bases, $f_1,...,f_m$, the test function $g$ is defined as a linear combination of the bases $g(Z) = v^T f(Z)$. The optimally weighted GMM (OWGMM) yields the minimum variance estimate of the linear weights $\theta$ by setting $R_f(f)=0$ and $R_g(v) = \frac{1}{4}v^T C v$ (see definition of the matrix $C$ in \cite{bennett2019deep} which is a function of the optimal weights $\tilde{\theta}$.}

\edit{\cite{bennett2019deep} extend OWGMM to the non-linear setting and parameterizes both $f$ ($Q$ in RL) and $g$ with neural networks,} $Q_\theta$ and $g_\phi$, and uses the following regularization,
\begin{align}
    R_Q(Q_\theta) &= 0\,, \nonumber\\
    R_g(g_\tau) &= \frac{1}{4} \expect[s,a\sim \mu_b, s'\sim P, r\sim R, a'\sim\pi]{g_\tau^2(s, a)
    (r - Q_{\tilde{\theta}}(s, a) + \gamma Q_{\tilde{\theta}}(s', a'))^2},
\end{align}
where $\tilde{\theta}$ should be a consistent estimator of the true parameter value. In practice, \cite{bennett2019deep} suggests to set $\tilde{\theta}$ to the latest estimate of $\theta$ in the iterative optimization process.

Deep GMM reduces to OWGMM in the linear function setting and results in the most efficient estimator in that case. The efficiency of this particular regularization scheme is not discussed in the non-linear case in \citet{bennett2019deep}. However, \citet[Sec.\ 6]{dikkala2020minimax} argues that such re-weighting is not required if one simply wants a fast convergence rate in the projected RMSE, defined next in \Cref{eq:projected_rmse}.

\subsubsection{Adversarial GMM Networks (AGMM)}
\citet{dikkala2020minimax} consider the general minimax objective function in \Cref{eq:gmm_objective} and focus on the generalization performance of the projected residual mean squared error, defined as
\begin{equation}
    \sqrt{\expect[Z]{\left(\expect[X]{\hat{f}(X) - f_0(X)|Z}\right)^2}} \,,
    \label{eq:projected_rmse}
\end{equation}
where $\hat{f}$ is the optimal solution of \Cref{eq:gmm_objective} on a dataset, and $f_0$ is the optimal solution of 
${\arg\inf}_{f\in\mathcal{F}} \sup_{g\in\mathcal{G}} \Psi(f, g)$.

\edit{The authors discuss the choice of the function spaces and regularization constants in order to derive a bound on the estimation error rate. They also instantiate the objective in different function spaces, including Reproducing Kernel Hilbert Spaces, High-dimensional Sparse Linear Function Spaces, Neural Networks, etc. When applying their theoretical findings to neural networks,} the regularizers on the function $Q_\theta(s, a)$ and $g_\tau(s, a)$ in the RL context are as follows\footnote{Personal communication with the authors},
\begin{align}
    R_Q(Q_\theta) &= a \|\theta\|_2^2\,, \nonumber\\
    R_g(g_\tau) &= b \|\tau\|_2^2 + \expect[s,a\sim \mu_b]{g_\tau^2(s, a)}\,,
\end{align}
where $a$ and $b$ are hyper-parameters for the $L_2$ regularization on the network parameters.

\subsubsection{Adversarial Structural Equation Models (ASEM)}
\edit{\cite{liao2020} introduce the generalized structural equation model (SEM) problem with IV being a special instance,
\begin{equation}
    A f = b,
\end{equation}
where $A: \mathcal{H} \rightarrow \mathcal{E}$ is a conditional expectation operator between two separable Hilbert spaces of square integrable functions, $f \in \mathcal{H}$ is the structural function of interest and $b\in \mathcal{E}$ is known or can be estimated.}
When applied to IV regression, this reduces to the same conditional moment restriction in \Cref{eq:moment_restriction},
under conditions $f\in L^2(\mathcal{X})$, $Af = \expect{f(X)|Z} \in L^2(\mathcal{Z})$, $b = \expect{Y|Z} \in L^2(\mathcal{Z})$.

\edit{With a similar dual formulation as other GMM methods introduced in preceding sections,} \cite{liao2020} propose a general solution to the SEM problem using an adversarial training approach with a minimax objective function, and both functions are parameterized with neural networks:

\edit{
\begin{equation}
    \label{eq:asem_objective}
    f^* = 
    \underset{f\in L^2(\mathcal{X})}{\arg\min}
    \max_{g\in L^2(\mathcal{Z})} \mathbb{E}[(f(X) - b(Z)) g(Z)] + \frac{\alpha}{2}\mathbb{E}[f(X)^2] - \frac{1}{2}\mathbb{E}[g(Z)^2]\,.
\end{equation}
}

They establish the consistency of the estimator theoretically under regularity conditions. In the RL context, the two networks should satisfy $Q_\theta \in L^2(\mathcal{S}\times \mathcal{A})$, $g_\tau \in L^2(\mathcal{S}\times \mathcal{A})$, and the corresponding regularizers are
\begin{equation}
    R_Q(Q_\theta) = \frac{\alpha}{2} \expect[s,a\sim \mu_b]{Q_\theta^2(s, a)},\qquad
    R_g(g_\tau) = \frac{1}{2} \expect[s,a\sim \mu_b]{g_\tau^2(s, a)},
\end{equation}
where $\alpha$ is a hyper-parameter. In this paper, we consider an additional $L_2$ regularization on the network parameters $\theta$ and $\tau$ as in AGMM, and tune the associated hyper-parameters together with $\alpha$. It is easy to see that AGMM described in the previous subsection is a special case of ASEM with $\alpha = 0$ (the different multiplier in $R_g$ does not change the solution of $Q_\theta$ after rescaling $g_\tau$ accordingly).

\subsubsection{Other adversarial IV methods}
\cite{lewis2018adversarial} propose another AGMM algorithm that minimizes the $L_2$ norm of the vector $$(\Phi_n^{(1)}, \Phi_n^{(2)}, \dots, \Phi_n^{(m)})$$ consisting of a finite set of moments (test functions), and solve it with a no-regret on line learning algorithm in an adversarial training fashion. This corresponds to a finite set of $\mathcal{G}$ and no regularizations in the objective of \Cref{eq:gmm_objective}.

\cite{muandet2019dual} consider a slightly different conditional moment objective,
\begin{equation}
    f^* = \underset{f\in\mathcal{F}}{\arg\min}~ \expect[Y,Z]{\left(Y - \expect[X|Z]{X|Z}\right)^2}\,.
\end{equation}
Note that the expectation with respect to $Y$ is outside of the square operator compared to \Cref{eq:moment_restriction}. This objective leads to a different dual formulation,
\begin{equation}
    \label{eq:dual_iv_objective}
    f^* = 
    \underset{f\in\mathcal{F}}{\arg\inf}
    \sup_{g\in\mathcal{G}} \Psi_n(f, g) - \frac{1}{2} \expect[Y,Z]{g(Y, Z)} \,,
\end{equation}
where the adversarial function $g$ is defined in the joint domain of $Y$ and $Z$. \cite{muandet2019dual} assume both $f$ and $g$ lie in reproducing kernel Hilbert spaces, which allows to obtain an analytical expression for the solution.

\subsubsection{Related OPE Methods}

Dual formulations have also been considered in the OPE literature, see e.g. \citep{nachum2019dualdice,yang2020off,mousavi2020black,uehara2020minimax}. Most of these works apply a saddle-point optimization method to estimate the density ratio between the distributions of state-action pairs under the evaluation policy $\pi$ and the behavioural policy $\pi_b$, each such distribution corresponding to the probability of encountering a state-action pair and averaging over time using the discount factor $\gamma$. This is an alternative approach to OPE based on importance sampling rather than the value function. 
\citet{yang2020off} point out that estimating the density ratio function is the dual formulation of estimating the $Q$ function with their particular objective. The closest work to our IV interpretation is \citet{uehara2020minimax}. The objective of their MQL algorithm is the squared $\Psi_n$ without regularization,
$$
    Q^* = 
    \underset{Q\in\mathcal{Q}}{\arg\inf}
    \sup_{g\in\mathcal{G}} \Psi_n^2(f, g)\,.
$$

\section{Experiments}\label{Sec:experiments}
In this section, we first demonstrate the advantage of IV methods over the biased DBRM and \edit{iterative Fitted Q Evaluation (FQE)} algorithm in the OPE problem using a simple MDP environment, and conduct an ablation study to investigate the influence of the model and algorithm parameters. We then propose a set of OPE benchmark problems with a varying level of randomness in the system dynamics. We evaluate the performance of all the non-linear IV methods in the paper on those problems, and compare to DBRM, FQE and distributional FQE which are state-of-the-art OPE method \citep{fu2021benchmarks}.

\subsection{Simple MDP problem}
\label{Sec:toy_experiment}
\begin{figure}[tbph!]
    \centering
    \includegraphics[width=\textwidth]{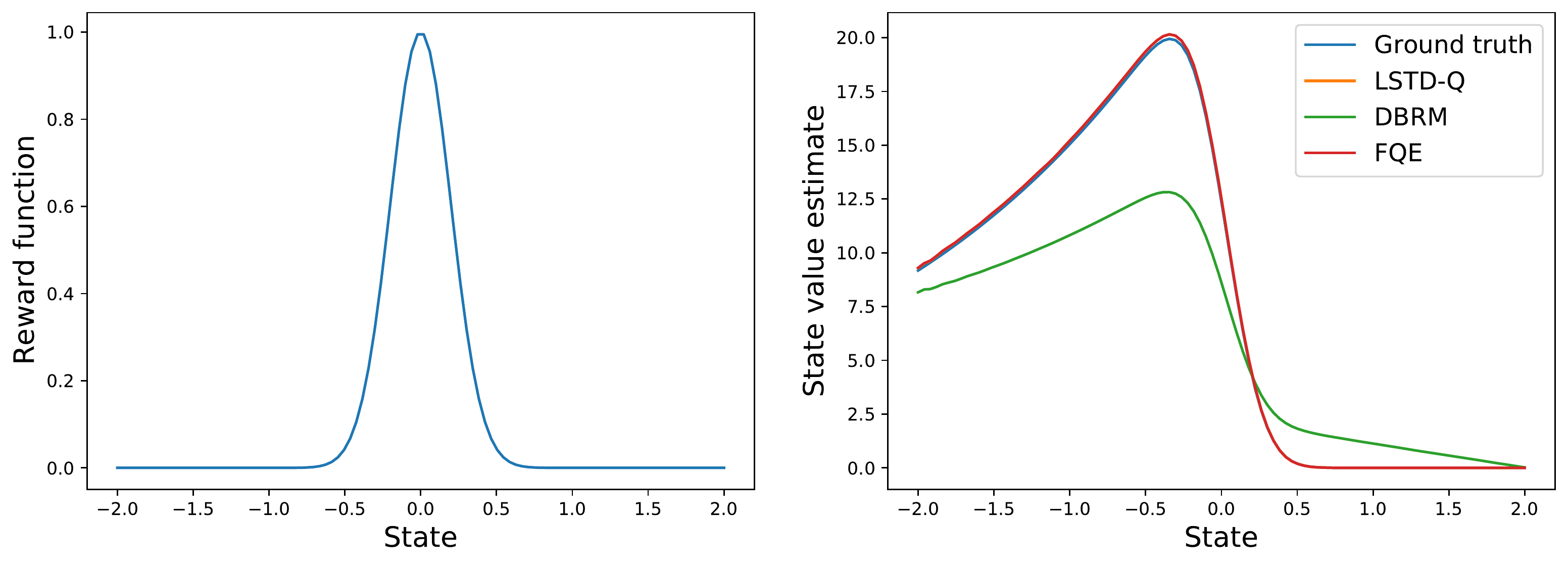}
        \caption{\edit{Simple MDP. Left: reward function. Right: The ground truth and estimated $Q(s, a=\text{right})$ by LSTD-Q, DBRM and FQE with $p=0.5$. LSTD-Q overlaps FQE.}}
        \label{fig:toy_exp_prob}
\end{figure}

Let us consider a simple MDP with 100 discrete states allocated uniformly along the interval $[-2, 2]$: $s_i = -2 + \frac{4}{100} i, i\in\{0, 1, \dots, 99\}$. The agent always starts at the first state $s_0$ in every episode and terminates at the last state $s_{99}$. There is only a single action, $a=\text{right}$, in every state to move to right, and therefore the policy is always fixed. The state is transitioned to the right neighboring state with a probability of $p > 0$ or stays in the same location. It is easy to show that the resulting state distribution $\mu(s)$ pooled from different steps across the trajectory, i.e.\ the offline data distribution, is uniform among all the non-terminating states $i\in\{0,1,\dots,98\}$, whatever the value of $p$ and $\mu(s_{99})=p \mu(s_{98})$. The reward function is defined with a Gaussian kernel as $R=\exp(-\frac{s^2}{0.2^2})$, and is illustrated in \Cref{fig:toy_exp_prob} together with $Q(s, a=\text{right})$. 

As there is only a single policy in this environment, the target policy is the same as the behavior policy. We sample a dataset of $N=10^5$ transitions with $p=0.5$, and estimate the state value using a fixed set of $D=90$ Gaussian kernel features $\phi_j(s) = \exp\left(\frac{(s-(-2+(4/D)j))^2}{0.1^2}\right),j=\{0,1,\dots,89\}$. For this linear instrumental variable regression problem, we compare the LSTD-Q method in \Cref{Sec:LSTD} with DBRM, which reduces to a naive least square minimization algorithm in this case, \edit{and FQE, which alternates solving a least squared minimization in \Cref{eq:fqe_linear} with the linear weights $\theta$ in $Y$ being fixed and replacing those weights with the solution}.

\begin{figure}[tbph!]
    \centering
    \includegraphics[width=\textwidth]{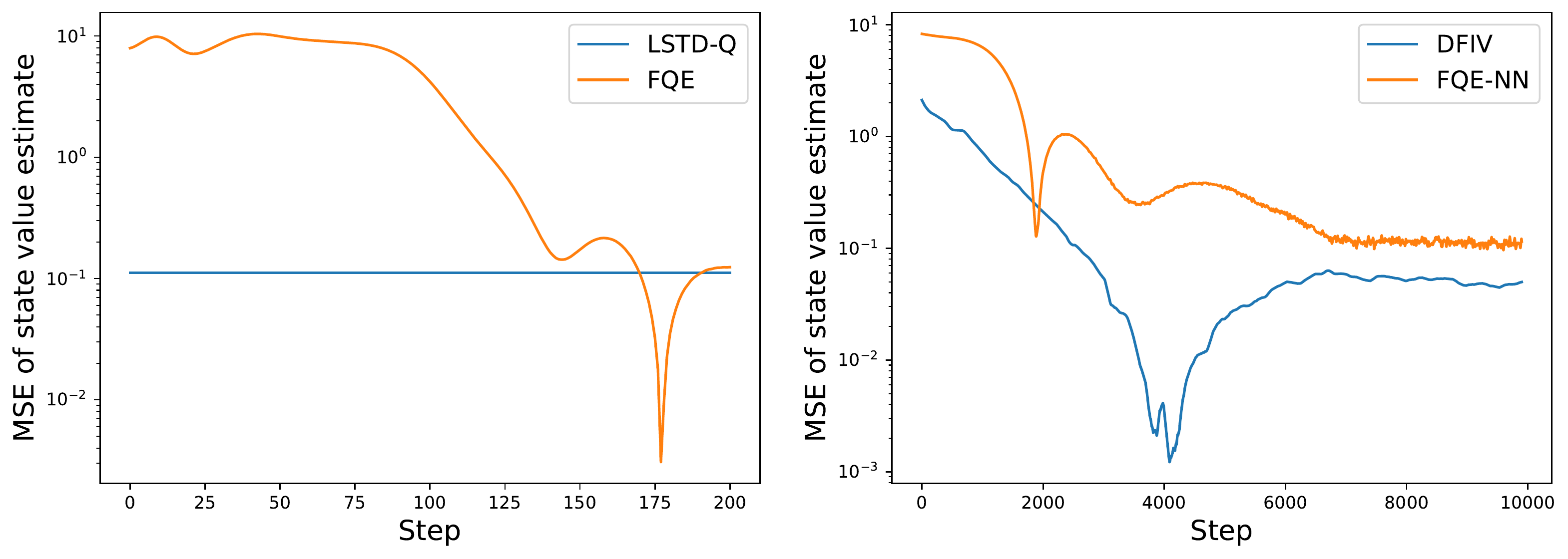}
    \caption{\edit{The absolute error of $Q(s_0, a=\text{right})$ as a function of optimization iteration in the linear (left) and non-linear setting (right). LSTD-Q is solved analytically in the linear setting.}}
    \label{fig:toy_exp_fqe}
\end{figure}

The estimates of the $Q$ function for all methods are shown in \Cref{fig:toy_exp_prob}. The estimate of LSTD-Q matches the ground truth value very well, while DBRM - which ignores the confounding problem - leads to a heavily biased estimate as expected. \edit{The estimate of FQE also matches the ground truth, but requires multiple iterations to converge to the solution as shown in \Cref{fig:toy_exp_fqe} (right). This is because the TD formulation of FQE can only propagate the state value information along the reverse order of state dynamics $s' \rightarrow s$ by one step at every iteration.}

Next, we conduct an ablation study to investigate how the advantage of IV method depends on the following four variables: dataset size $N$, feature dimensions $D$, transition randomness ($1-p$), and the extent of off-policyness. While the target policy always matches the behavior policy in this problem, we create an offline dataset with a shifted distribution by sampling the states with the following distribution $\mu(s) \propto \exp(\alpha s)$, where $\alpha=0$ corresponds to the original uniform distribution, and a larger value of $\alpha$ leads to a shifted distribution towards the right end of the state space.

\begin{figure}[tbph!]
    \centering
    \includegraphics[width=\textwidth]{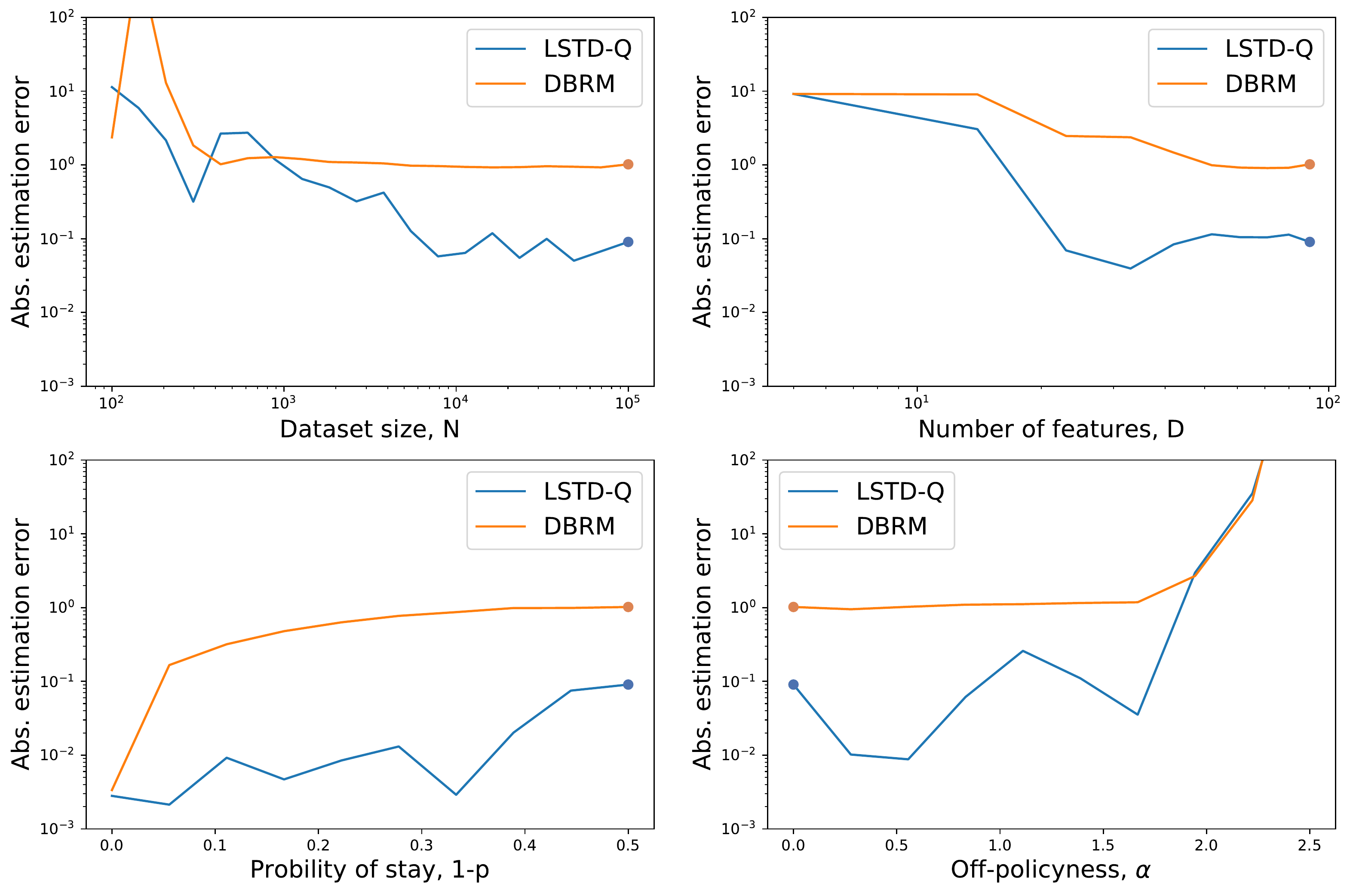}
        \caption{Simple MDP ablation study. Each plot shows the absolute error of $Q(s_0, a=\text{right})$ as a function of dataset size, number of features, stochasticity of the dynamics, and the distribution shift between the dataset and that generated by the target policy. The dots represent the default setting in \Cref{fig:toy_exp_prob}.}
        \label{fig:toy_exp_ablation}
\end{figure}

We display the absolute error of the estimated state value at the initial state $Q(s_0, a=right)$ {by LSTD-Q and DBRM} in \Cref{fig:toy_exp_ablation}. \edit{The error of FQE is not shown because it converges to the same solution as LSTD-Q in the linear case.} We find the error of LSTD-Q increases and eventually becomes on par with DBRM when we reduce the size of the dataset, or increase the data distribution shift, which also decreases the effective dataset size. The error of LSTD-Q also increases when we reduce the number of features, which will lead to model misspecification, and could violate the IV requirement that the residual needs to have zero mean. Lastly, we see that DBRM is unbiased when the dynamics are deterministic, $p = 1$, but the bias increases quickly when $p$ decreases. In contrast, the error of LSTD-Q increases but much more slowly, which we suspect is due to an increasingly diverse distribution of transitions $(s,s')$ that requires more data to learn the estimate accurately.

\edit{When we use a non-linear value function, the solutions given from IV techniques do not coincides with FQE, and both methods require iterative optimization. Nonetheless, the policy evaluation algorithms based on non-linear IV (e.g. DFIV) preserve the relatively faster convergence rate than FQE in this example, as shown in \Cref{fig:toy_exp_fqe} (right). Here we estimate the value function from the raw state $s$ and estimate an multi-layer perceptron (MLP) with two hidden layers, each with 50 units and ReLU activation function. Both methods use a learning rate of $10^{-4}$.}

From this ablation study, we demonstrate that the advantages of using an IV method over a simple method like DBRM and FQE can be significant, but that the magnitude of this benefit depends on multiple variables. On a more complex and comprehensive OPE benchmark, we study in the following section whether recent non-linear IV methods are competitive with state-of-the-art OPE methods such as FQE and Distributional FQE \citep{fu2021benchmarks}. 

\subsection{OPE benchmark problems}
\subsubsection{Environments}
\begin{figure}[tbph!]
     \centering
     \begin{subfigure}[b]{0.3\textwidth}
         \centering
         \includegraphics[width=\textwidth]{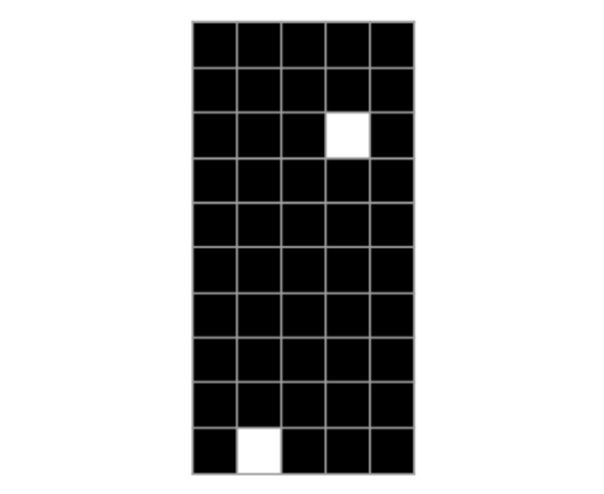}
         \caption{Catch}
         \label{fig:bsuite_catch}
     \end{subfigure}
     \hfill
     \begin{subfigure}[b]{0.3\textwidth}
         \centering
         \includegraphics[width=\textwidth]{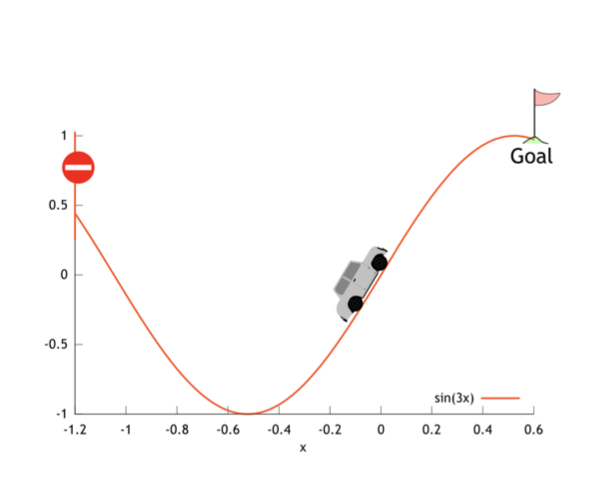}
         \caption{Mountain Car}
         \label{fig:bsuite_mountain_car}
     \end{subfigure}
     \hfill
     \begin{subfigure}[b]{0.3\textwidth}
         \centering
         \includegraphics[width=\textwidth]{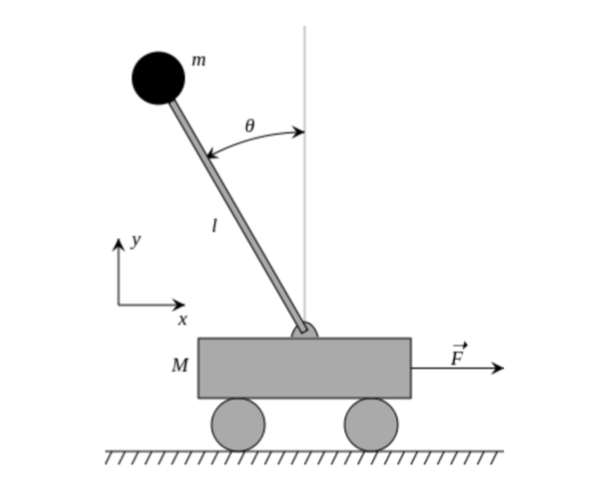}
         \caption{Cartpole}
         \label{fig:bsuite_cartpole}
     \end{subfigure}
        \caption{Three BSuite tasks}
        \label{fig:bsuite}
\end{figure}

\begin{figure}[tbph!]
     \centering
     \begin{subfigure}[b]{0.23\textwidth}
         \centering
         \includegraphics[width=\textwidth]{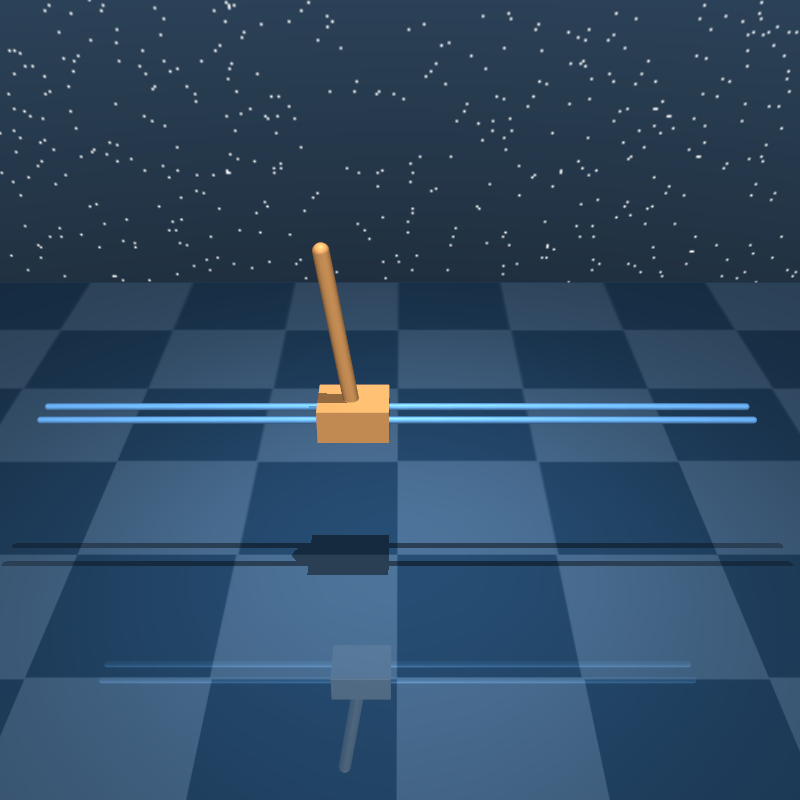}
         \caption{Cartpole Swingup}
         \label{fig:control_cartpole}
     \end{subfigure}
     \hfill
     \begin{subfigure}[b]{0.23\textwidth}
         \centering
         \includegraphics[width=\textwidth]{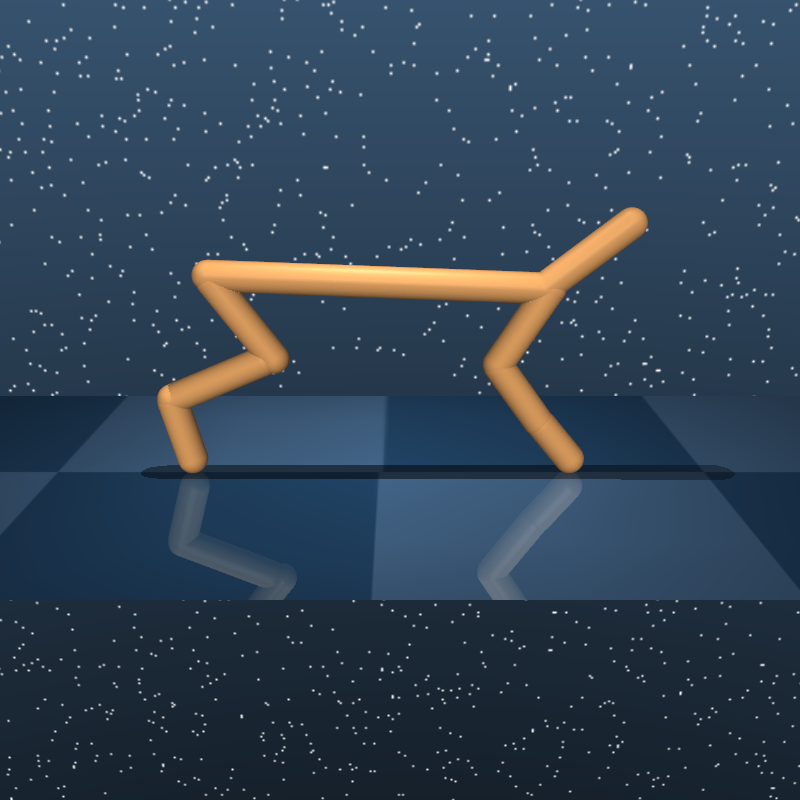}
         \caption{Cheetah Run}
         \label{fig:control_cheetah}
     \end{subfigure}
     \hfill
     \begin{subfigure}[b]{0.23\textwidth}
         \centering
         \includegraphics[width=\textwidth]{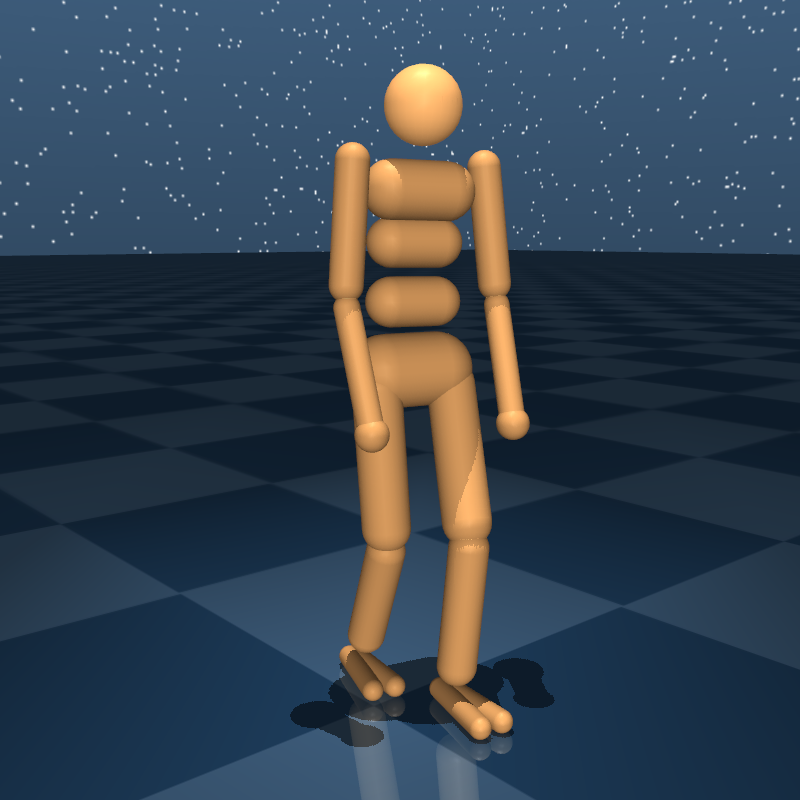}
         \caption{Humanoid Run}
         \label{fig:control_humanoid}
     \end{subfigure}
     \hfill
     \begin{subfigure}[b]{0.23\textwidth}
         \centering
         \includegraphics[width=\textwidth]{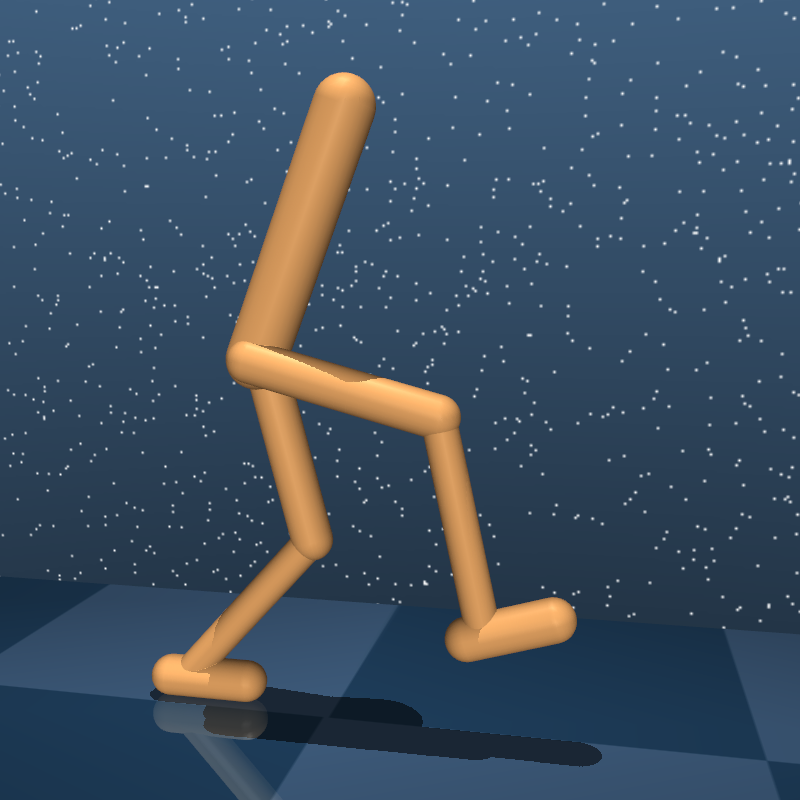}
         \caption{Walker Walk}
         \label{fig:control_walker}
     \end{subfigure}
        \caption{Four DM Control tasks}
        \label{fig:dm_control}
\end{figure}

We consider a list reinforcement learning environments from two widely used task collections: Behaviour Suite (BSuite) \citep{osband2019behaviour} and DeepMind Control Suite (DM Control) \citep{tassa2020dmcontrol}. BSuite is a collection of traditional RL environments with a discrete action space. We choose three environments that can be solved by a standard DQN agent \citep{mnih2015humanlevel}: Catch, Cartpole, and Mountain Car. DM Control is a collection of physics-based simulation environments, using MuJoCo physics, for studying continuous control problems with a continuous action space. We choose four environments that can be solved by a standard D4PG agent \citep{barth2018distributed}: Cartpole Swingup, Cheetah Run, Walker Walk, Humanoid run. A brief description of each of the seven environments is provided as follows with illustrations in \Cref{fig:bsuite,fig:dm_control}:

\begin{itemize}
    \item BSuite
    \begin{itemize}
        \item Catch: A 10x5 Tetris-grid with single block falling per column. The agent can move left/right in the bottom row to ‘catch’ the block.
        \item Mountain Car: The agent drives an underpowered car up a hill~\citep{moore1990efficient}.
        \item Cartpole: The agent can move a cart left/right on a plane to keep a balanced pole upright~\citep{barto1983neuronlike}.
    \end{itemize}
    \item DM Control
    \begin{itemize}
        \item Cartpole Swingup: Swing up and balance an unactuated pole by applying forces to a cart at its base. The physical model conforms to \citet{barto1983neuronlike}.
        \item Cheetah Run: A running planar biped based on \citet{wawrzynski2009real}.
        \item Humanoid Run: A simplified humanoid with 21 joints, based on the model in \citet{tassa2012synthesis}.
        \item Walker Walk: An improved planar walker based on the one introduced in \citet{lillicrap2015continuous}.
    \end{itemize}
\end{itemize}

Every environment has a real-valued state space $\mathcal{S}\subseteq \mathcal{R}^{D_S}$, and a discrete action space $\mathcal{A}=\{0, 1, 2, \dots, D_A-1\}$ for BSuite tasks and continuous action space $\mathcal{A}\subseteq \mathcal{R}^{D_A}$ for DM Control tasks, respectively.

All environments have deterministic system dynamics, which would be hard to find in real-world applications. In order to study how IV methods can address the bias introduced through the confounding variable of the next state $s'$, we modify the original environments with additional randomness in the dynamics. Specifically, for BSuite environments, we randomly replace the agent action by a uniformly sampled action with a probability of $p\in \{0, 0.1, 0.2, 0.3, 0.4, 0.5\}$, resulting in a stochastic transition distribution,
\begin{equation}
    \tilde{P}(s'|s, a) = (1 - p) P(s'|s, a) + \frac{p}{D_A} \sum_{\tilde{a}=0}^{D_A} P(s'|s, \tilde{a}) \,.
\end{equation}
For DM Control environments, we insert some Gaussian noise to the agent action, $\varepsilon \sim \mathcal{N}(0, \sigma^2)$ with $\sigma \in \{0, 0.2, 0.4, 0.6, 0.8, 1.0\}$, resulting in a stochastic transition distribution,
\begin{equation}
    \tilde{P}(s'|s, a) = \int_{\varepsilon} P(s'|s, a + \varepsilon) \mathcal{N}(\varepsilon|0, \sigma^2)\mathrm{d}\varepsilon \,.
\end{equation}
Note that the randomly perturbed action is transparent to the agent. When $p=0$ or $\sigma=0$, it reduces to the original deterministic environment.

Overall, we create 42 environments for our experiments with 7 different tasks and 6 levels of dynamics randomness per task.
The state and action dimensions are provided in Table~\ref{tab:bsuite} and \ref{tab:dm_control}.

\begin{table}[tbh!]
    \centering
    \begin{tabular}{p{3cm}|l|ccc}
        & & \textbf{Catch} & \textbf{Mountain Car} & \textbf{Cartpole} \\
        \hline
        \multirow{2}{*}{\textbf{Dimensions}} 
        & $D$ & 50 & 3 & 6\\
        & $A$ & 3 & 3 & 3\\
        \hline
        \multirow{1}{*}{\textbf{Target policy}}
        & Train Episodes     & 2K   & 500             & 1K\\
        \hline
        \multirow{4}{3cm}{\textbf{Near-policy}\newline \textbf{Dataset}}
        & Train  Episodes     & 20K   & 5K                & 1K\\
        & Train  Transitions  & 180K  & 759K$\sim$1.59M  & 325K$\sim$833K\\
        & Valid  Episodes     & 2K    & 500               & 100\\
        & Valid  Transitions  & 19K   & 75.1K$\sim$158K     & 33K$\sim$83K\\
        \hline
        \multirow{4}{3cm}{\textbf{Pure Offline}\newline \textbf{Dataset}}
        & Train  Episodes     & 1.8K   & 450            & 900\\
        & Train  Transitions  & 16.2K  & 75K$\sim$160K  & 512K$\sim$671K\\
        & Valid  Episodes     & 200    & 50             & 100\\
        & Valid  Transitions  & 1.8K   & 8K$\sim$14K    & 71K$\sim$80K\\
        \hline
    \end{tabular}
    \caption{BSuite tasks. Every Catch episode has 9 transitions. The average length of an episode in the Mountain Car/Cartpole increases/decreases as the level of environment randomness $p$ increases. The training and validation data ratio is 9:1.}
    \label{tab:bsuite}
\end{table}

\begin{table}[tbh!]
    \centering
    \begin{tabular}{p{3cm}|l|cccc}
        & & \textbf{{Cartpole}} & \textbf{Cheetah} & \textbf{Humanoid} & \textbf{Walker} \\
        & & \textbf{{Swingup}} & \textbf{Run} & \textbf{Run} & \textbf{Walk} \\
        \hline
        \multirow{2}{*}{\textbf{Dimensions}} 
        & $D$ & 5 & 17 & 67 & 24\\
        & $A$ & 1 & 6 & 21 & 6\\
        \hline
        \multirow{1}{*}{\textbf{Target policy}}
        & Train Episodes     & 300   & 4K    & 100K  & 1.5K\\
        \hline
        \multirow{4}{3cm}{\textbf{Pure Offline}\newline \textbf{Dataset}}
        & Train  Episodes     & 270   & 3.6K    & 9K    & 1.35K\\
        & Train  Transitions  & 270K  & 3.6M    & 9M    & 1.35M\\
        & Valid  Episodes     & 30    & 400     & 1K    & 150\\
        & Valid  Transitions  & 30K   & 400K    & 1M    & 150K\\
        \hline
    \end{tabular}
    \caption{DM Control Suite tasks. Every episode has 1000 transitions. The training and validation data ratio is 9:1. The offline dataset of the Humanoid Run task is subsampled with by 10\% from the 100K episodes generated from the training process.}
    \label{tab:dm_control}
\end{table}

\subsubsection{Target policies for evaluation and offline datasets}
We run the default DQN agent for every random level of the three BSuite tasks and the default D4PG agent for the four DM Control tasks from the ACME library \citep{hoffman2020acme} until the episodic return does not increase noticeably any more. The number of training episodes is provided in Table~\ref{tab:bsuite} and \ref{tab:dm_control}. We use the learned policy with a small amount of action noise as the target policy for evaluation. For the DQN agent, we use an $\varepsilon$-greedy policy, that is, taking the greedy action of $\argmax_a Q(s, a)$ with a probability $1-\varepsilon$ and a random action otherwise where $\varepsilon=0.1$. For the D4PG agent, we use the learned policy network with an additive Gaussian noise with a standard deviation of $0.2$.

We consider two types of offline datasets with a different level of difficulty for OPE: an easy near-policy dataset and a hard pure offline dataset. The size of the dataset for each environment is given in Table~\ref{tab:bsuite} and \ref{tab:dm_control}.

For the easy dataset, we consider the three BSuite tasks only. For every task we define the behavior policy in the same way as the target policy except with a slightly larger exploration probability of $\varepsilon=0.3$ instead of $0.1$. Therefore the behavior policy is close to the target policy. We then play the behavior policy in the corresponding environment repeatedly and collect a sufficiently large off-policy dataset for each task.

For the hard dataset, we restart the agent training process with a different random seed and collect the episodes along the training. The resulting dataset consists of episodes generated from various partially trained policies, some of which are close to the initial random policies while others are close to a well-optimized policy. The dataset is then split randomly into training and validation subsets with a ratio of 9:1. This is akin to the data generation protocol in the RL Unplugged dataset \citep{gulcehre2020rl} with two differences: (1) we modify environments with random dynamics from the original deterministic environment, (2) the dataset is collected from the training process of a different random seed, therefore the policies used to generate the dataset could be substantially different from the target policy to be evaluated.

\subsection{Experiment setup and hyper-parameter selection}

We compare a list of representative non-linear IV methods, including Kernel IV (KIV), Deep IV, Deep Feature IV (DFIV) and three adversarial IV methods: Deep GMM, Adversarial GMM Networks (AGMM), Adversarial approach to structural equation models (ASEM). We also include as baselines the deterministic Bellman residual minimization (DBRM) and two variants of the fitted Q evaluation methods with a deterministic (FQE) and distributional (DFQE) Q representation respectively. (D)FQE was shown to be the best performing OPE algorithm in a recent benchmark paper \citep{fu2021benchmarks} under a similar task setting as in this paper.

All algorithms except KIV use the same network architecture to estimate the Q function as in the trained agent for a fair comparison.
For BSuite tasks, the $Q$ network is an MLP with layer size 50-50-1 and ReLU activation. The input is a concatenation of the flattened observation and one-hot encoding of the discrete action variable. 
For DM Control tasks, it is an MLP with layer size 512-512-256-1, ELU activation and a layer normalization after the first hidden layer. The input is the concatenation of the flattened observation and action variables.
The architecture of additional networks in each algorithm, such as the generative model in Deep IV and the adversarial function network in AGMM, ASEM and DeepGMM are selected as part of the hyper-parameter search procedure that will be discussed later. We use OAdam for adversarial methods as suggested by \citet{bennett2019deep,dikkala2020minimax} and Adam for other methods.

We compare all the algorithms with respect to the accuracy of estimating the target policy value $\rho(\pi)$ (Eq.~\ref{eq:policy_value}), i.e., the expected cumulative discounted reward from the initial state distribution. The estimate is computed as
\begin{equation}
\hat{\rho}(\pi)=\mathbb{E}_{s\sim \mu_0, a|s\sim \pi}[\hat{Q}^{\pi}(s,a)],
\end{equation}
where $\hat{Q}^{\pi}$ is given by each algorithm under comparison. We normalize the policy value into a range of $[0, 1]$ for ease of comparison across environments:
\begin{equation}
    \rho_{\mathrm{Norm}} = \frac{\rho - \rho_{\mathrm{min}}}{\rho - \rho_{\mathrm{max}}}
\end{equation}
where $\rho_{\min / \max} := -1 / 1$ for BSuite Catch and $\rho_{\min / \max} := R_{\min / \max}\sum_{t=0}^{1000}\gamma^t$ for other environments. We measure the accuracy in terms of the absolute error $|\hat{\rho}_{\mathrm{Norm}}-\rho_{\mathrm{Norm}}|$ in this paper. Other metrics such as the policy ranking correlation and regret have been considered in the literature \citep{paine2020hyperparameter,fu2021benchmarks} when multiple target policies are available in the same environment. Our experiment setup does not meet that condition and those metrics are hence not included.

Some of the algorithms implemented are sensitive to the choice of hyper-parameters. In order to ensure a fair comparison, we run a thorough hyper-parameter search for every algorithm in every environment. We randomly sample up to 100 hyper-parameter settings for every algorithm and choose the setting with the best metric on a held-out validation dataset. Due to the large number of tasks (environment and dataset combinations), we search for the best hyper-parameter at one environment random level in every dataset ($p=0.2$ for BSuite and $\sigma=0.4$ for DM Control tasks) and apply the same setting to other levels. Once the hyper-parameter is selected, we run each algorithm with 5 random seeds for every task to measure the mean and variance of the estimate.

Note that due to the state distribution shift between the behavior policy and target policy, the best hyper-parameter setting on the validation dataset from the behavior distribution does not guarantee a good performance when evaluating the target policy value. It remains an open research problem how to select the hyper-parameter for OPE given one does not have access to the ground truth value \citep{paine2020hyperparameter}. We explain the metric adopted for selecting the hyper-parameters of each algorithm in details in \Cref{Sec:hyper}.

\subsection{Results}
\subsubsection{Near-policy dataset}
\begin{figure}[tbhp!]
    \centering
    \includegraphics[width=\textwidth]{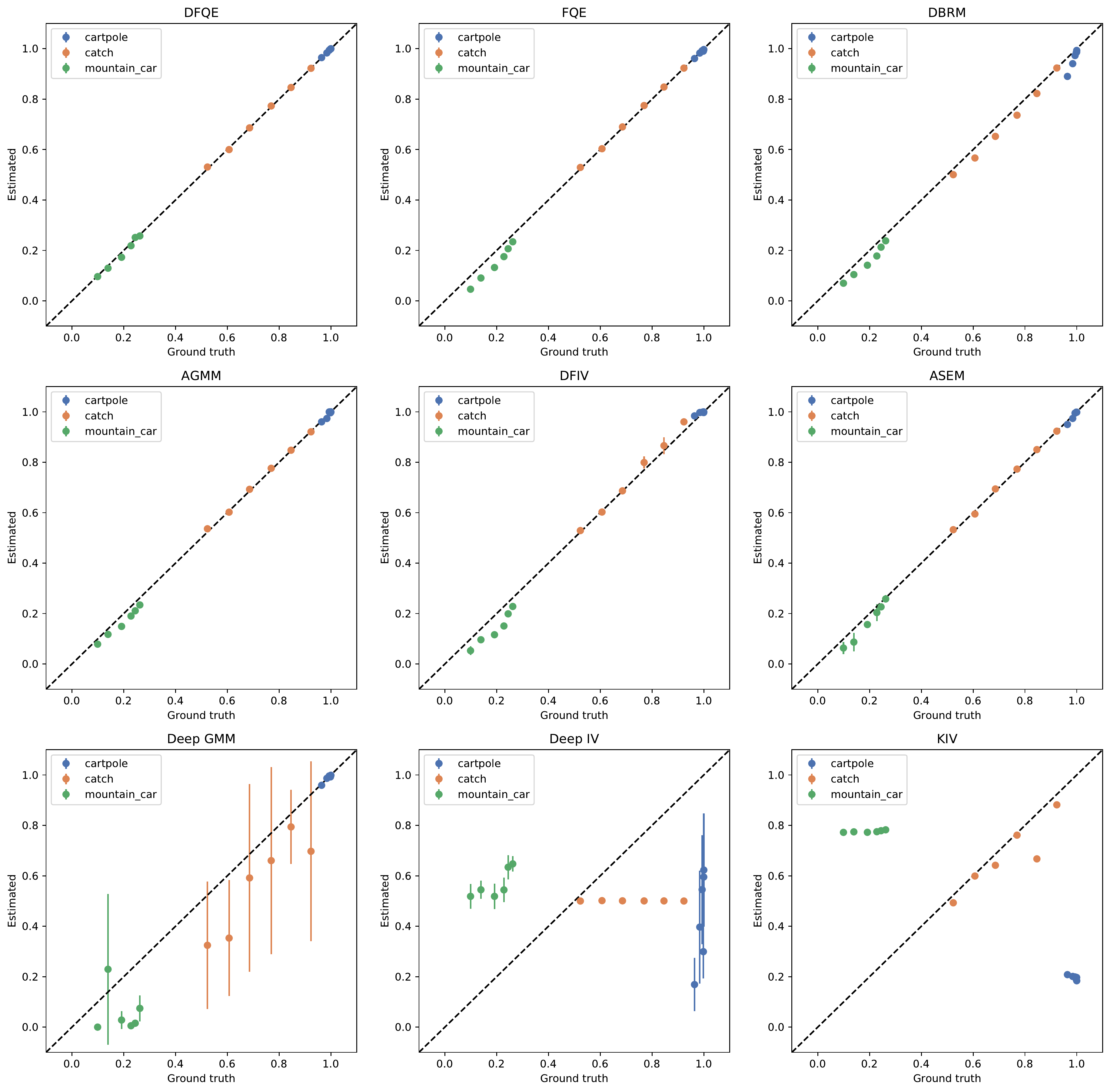}
    \caption{Estimated policy value vs groundtruth with the near-policy dataset}
    \label{fig:near_policy_scatter}
\end{figure}

\begin{figure}[tbhp!]
    \centering
    \includegraphics[width=\textwidth]{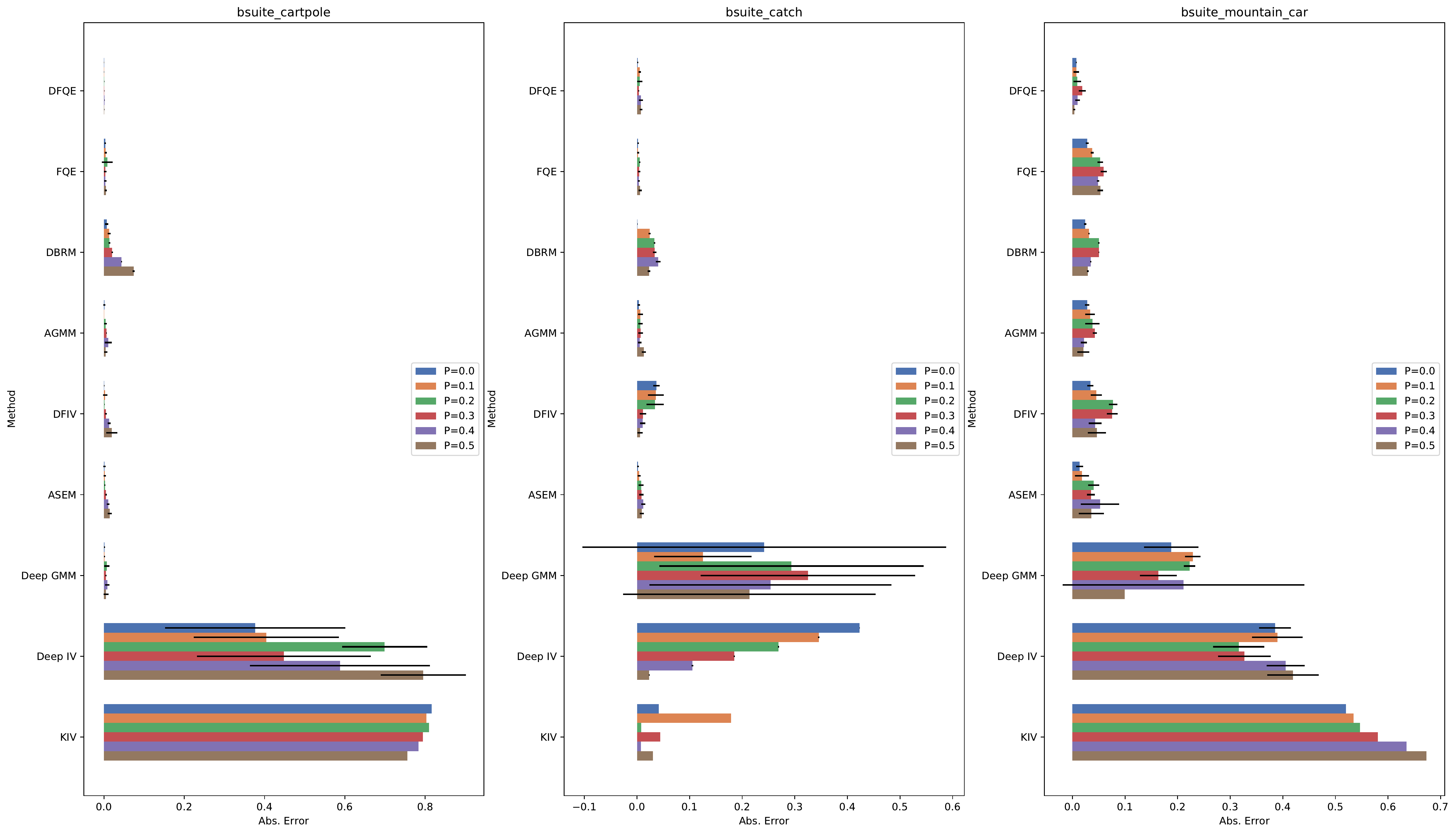}
    \caption{Absolute error of policy value estimation 
        with the near-policy dataset}
    \label{fig:near_policy_mae}
\end{figure}

\begin{figure}[tbhp!]
    \centering
    \includegraphics[width=0.6\textwidth]{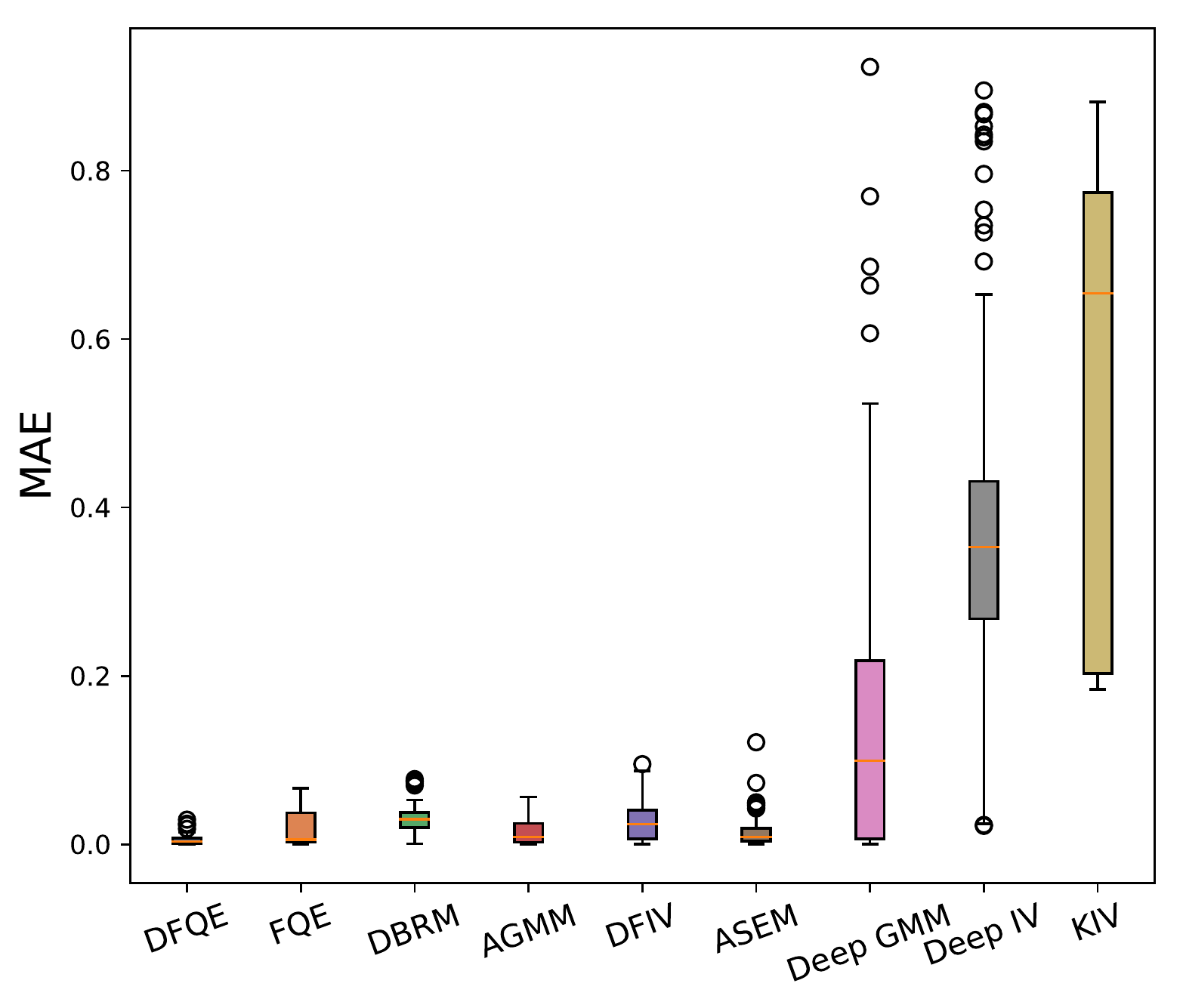}
    \caption{Distribution of the absolute error across all tasks with the near-policy dataset}
    \label{fig:near_policy_mae_pool}
\end{figure}

We first study the performance of all the algorithms on the easy offline dataset with a near-policy data distribution and sufficiently larget data size. \Cref{fig:near_policy_scatter} shows the scatter plot of the estimated policy value versus the ground-truth value. Each dot represents the mean and 1-standard deviation of the estimate from 5 random runs for every environment and every random level. Additionally, we show the absolute error of the estimates
for each task in \Cref{fig:near_policy_mae} and a box-plot of the distribution of errors pooled from all tasks as a summary in \Cref{fig:near_policy_mae_pool}.


\begin{figure}[tbhp!]
    \centering
    \includegraphics[width=0.45\textwidth]{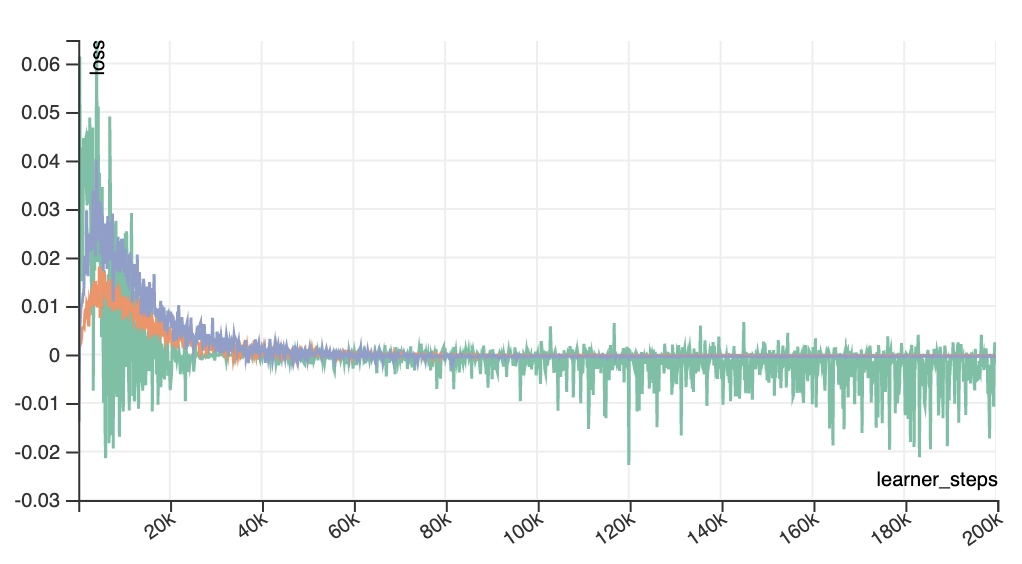}
    ~
    \includegraphics[width=0.45\textwidth]{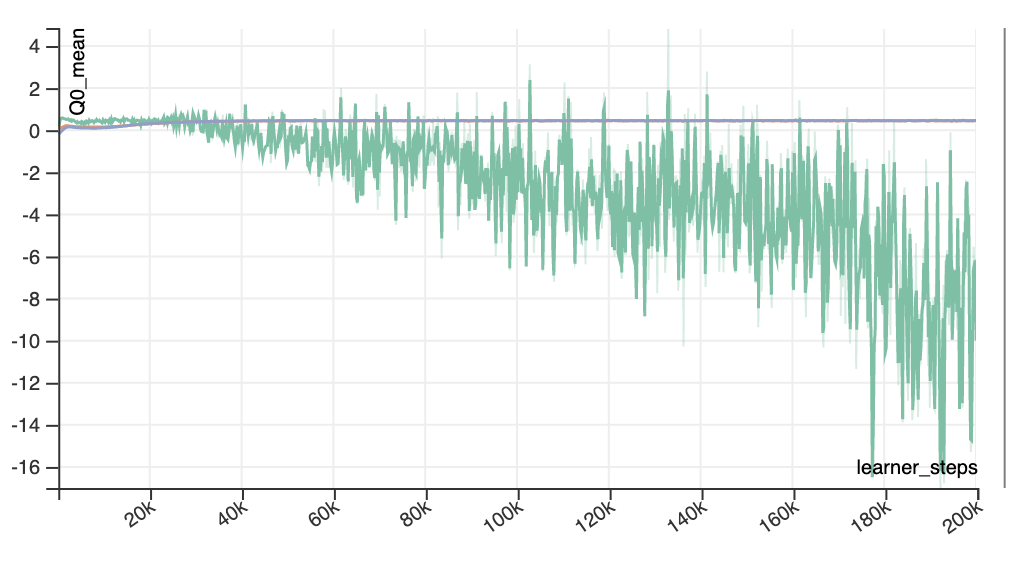}
    \caption{A typical trajectory of the training loss (left) and the estimated unnormalized policy value (right) of DeepGMM (green), AGMM (orange), and ASEM (blue) as a function of training steps. The plots are obtained from the task of BSuite Catch with environment random level $p=0.2$. A valid policy value should be in the range of $[-1, 1]$. The curves of AGMM and ASEM overlap on even other in the right plot.}
    \label{fig:deepgmm_curve}
\end{figure}

Most algorithms provide an accurate estimate of the policy value on this dataset, except Deep GMM, Deep IV, and KIV. We observe experimentally that the training process of Deep GMM is very unstable compared the other two adversarial approaches (AGMM and ASEM). \Cref{fig:deepgmm_curve} shows a typical trajectory of the training loss and the estimated policy value along the training process. We suspect it is due to the use of the optimal weighting $\mathcal{C}_{\tilde{\theta}}$ in the regularization term.
Deep IV fails with both a large mean absolute error and variance. In particular, in the Catch environment, the generative model completely fails to predict the next state. This illustrates the challenge of modeling a moderately-high dimensional state space (50-dimensions) using a simple feed-forward network to predict the parameters of a mixture of Gaussian generative model as proposed in \citet{hartford2017deep}. We expect the performance to improve with a more sophisticated generative model as evidenced by recent model-based OPE work in \citet{zhang2021autoregressive}.
KIV fails in the Cartpole and Mountain Car environments with a large error too. This is in agreement with the observation by \citet{xu2021learning} that shallow features are not capable of modeling complex structural functions.

\Cref{fig:near_policy_mae_pool} shows that DFQE is the most accurate OPE method on this dataset. Among the IV methods, AGMM, ASEM and DFIV all give fairly small estimation errors across all tasks while AGMM performs the best.

\subsubsection{Bias in DBRM}
The bias in DBRM due to ignoring the confounding variable $s'$ is clearly observed in the Cartpole and Catch environments where the error increases with the level of randomness.

\begin{figure}[tbhp!]
    \centering
    \includegraphics[width=0.6\textwidth]{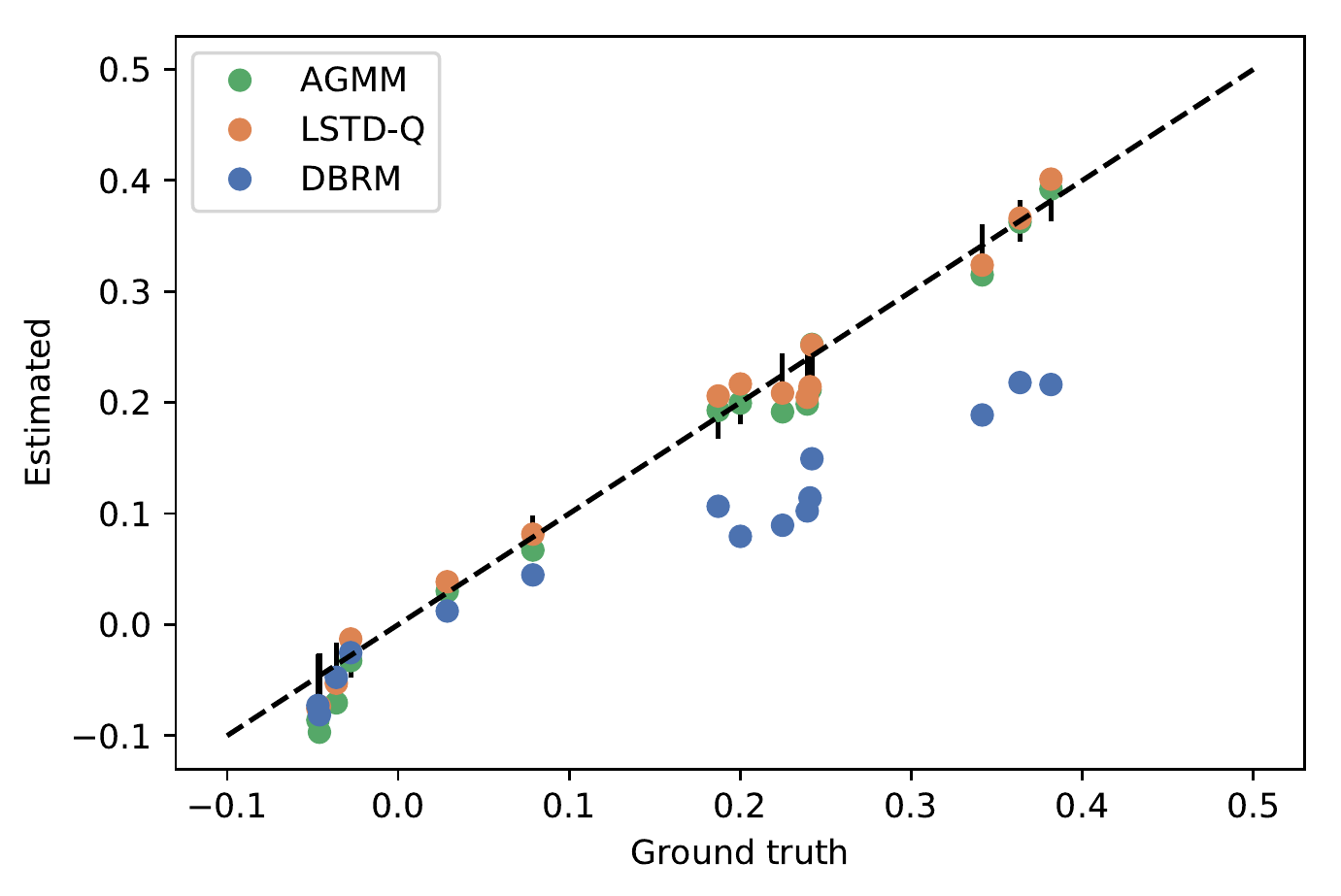}
    \caption{Estimated policy value in the DBRM ablation study. Every dot represents one initial state-action pair. The error bar shows the 95\% confidence interval of the ground-truth value estimation.}
    \label{fig:dbrm_ablation}
\end{figure}

We further inspect its bias in the $Q$ estimation in the following ablation study using the Catch environment with an environment random level $p=0.4$ as an example. After AGMM converges, we take the last hidden layer of the value network as a set of deep features, and re-fit the weights of the output linear layer. Given the fixed features, it reduces to a standard IV problem with linear functions. We fit the linear weights with two methods, DBRM and LSTD-Q, and compare the new estimated Q value of the 15 unique initial state-action pairs of Catch together with those obtained by AGMM in \Cref{fig:dbrm_ablation}. The estimates of LSTD-Q and AGMM match the ground truth very well within a 95\% confidence interval while DBRM estimate shows a significantly large error. This suggests that DBRM suffers, as expected, from a bias in the presence of environment randomness even when using a good set of deep features.

\subsubsection{Pure offline dataset}
We then evaluate all the algorithms on the hard dataset in both BSuite and DM Control environments. The episodes are generated from a mixture of partially trained policies from a different run, and the distribution of states is likely to have a quite different coverage from the distribution generated by the target distribution.

\begin{figure}[tbhp!]
    \centering
    \includegraphics[width=\textwidth]{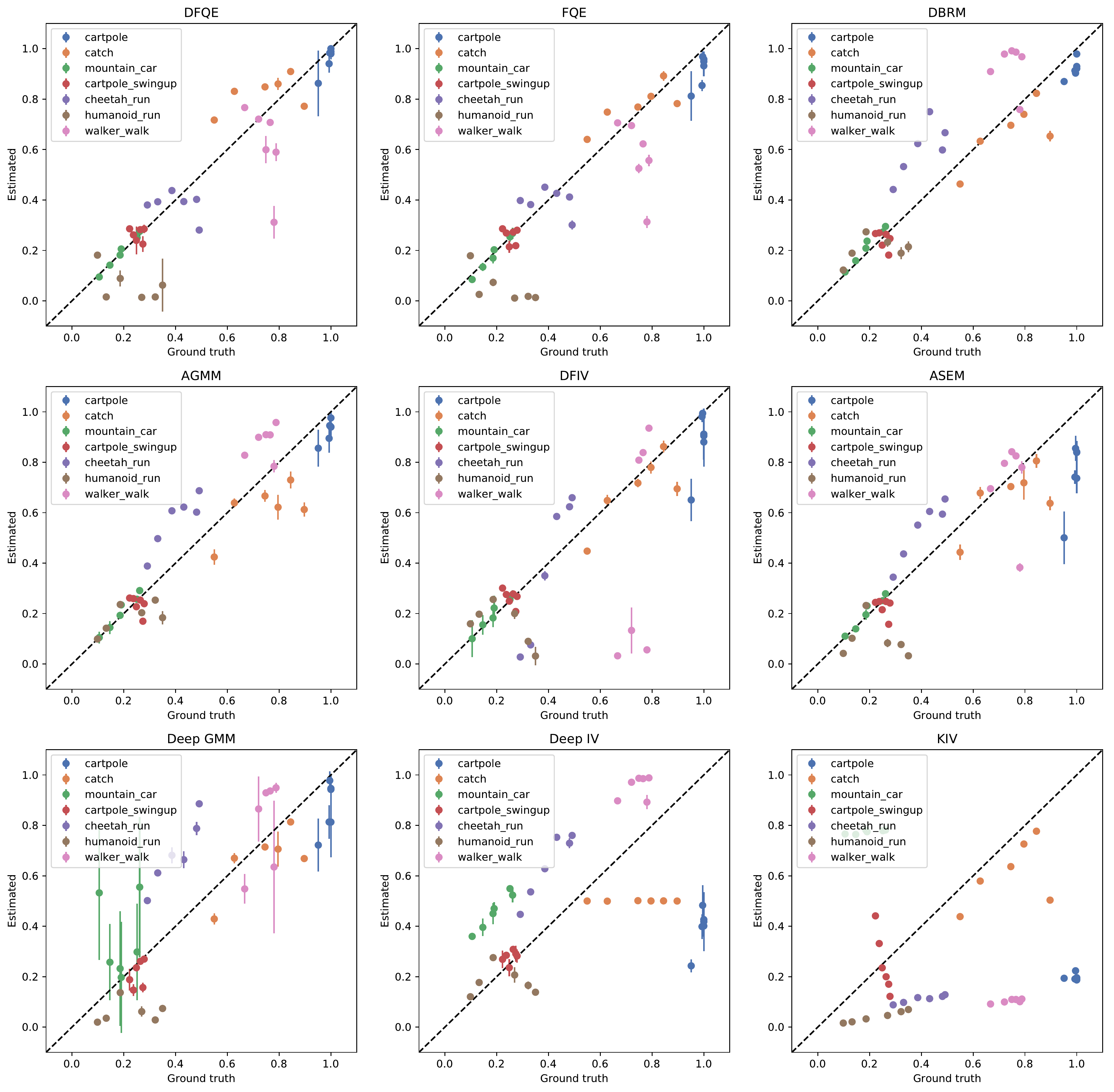}
    \caption{Estimated policy value vs groundtruth with the offline dataset}
    \label{fig:offline_policy_scatter}
\end{figure}

\begin{figure}[tbhp!]
    \centering
    \includegraphics[width=\textwidth]{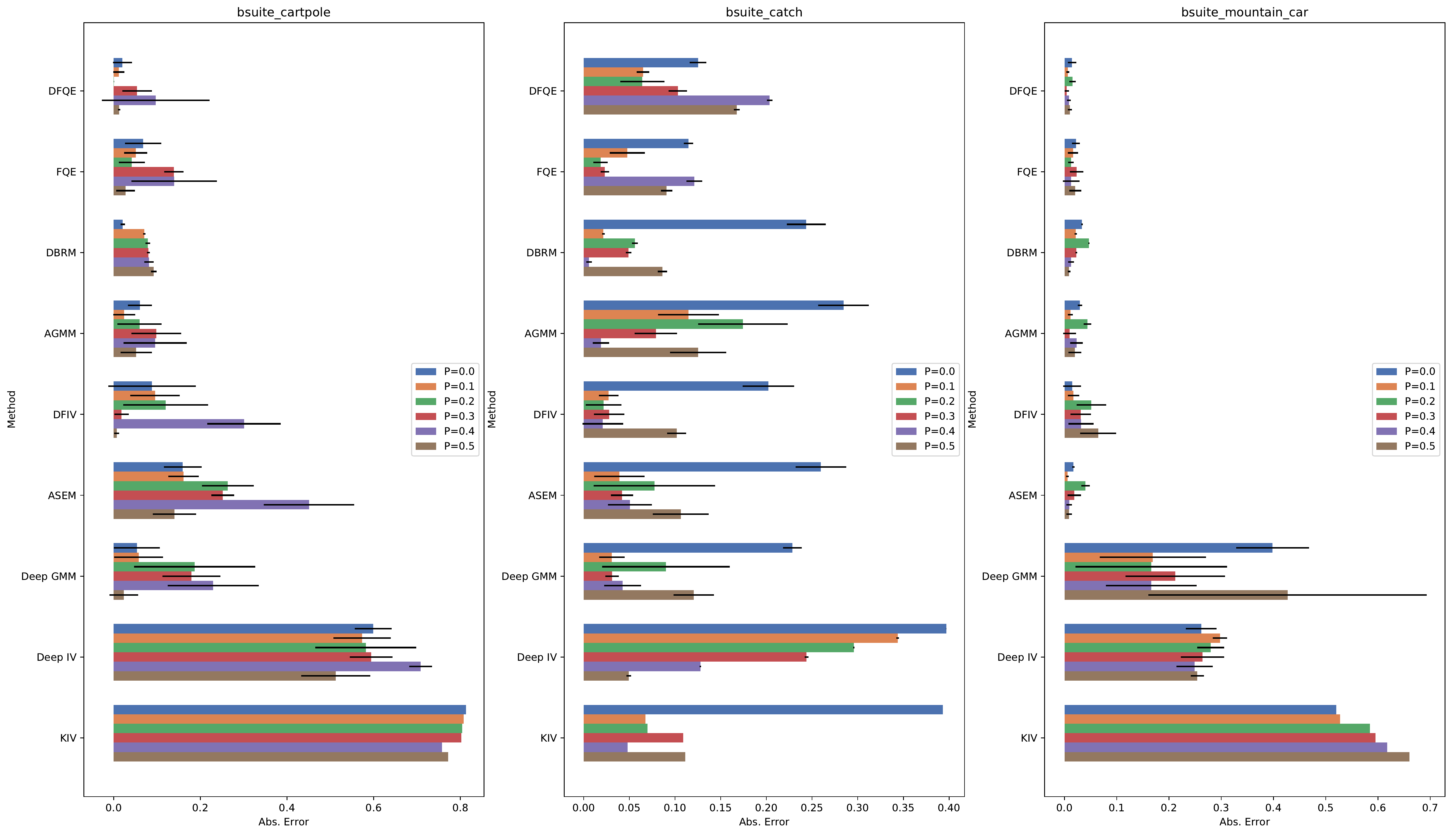}\\
    \includegraphics[width=\textwidth]{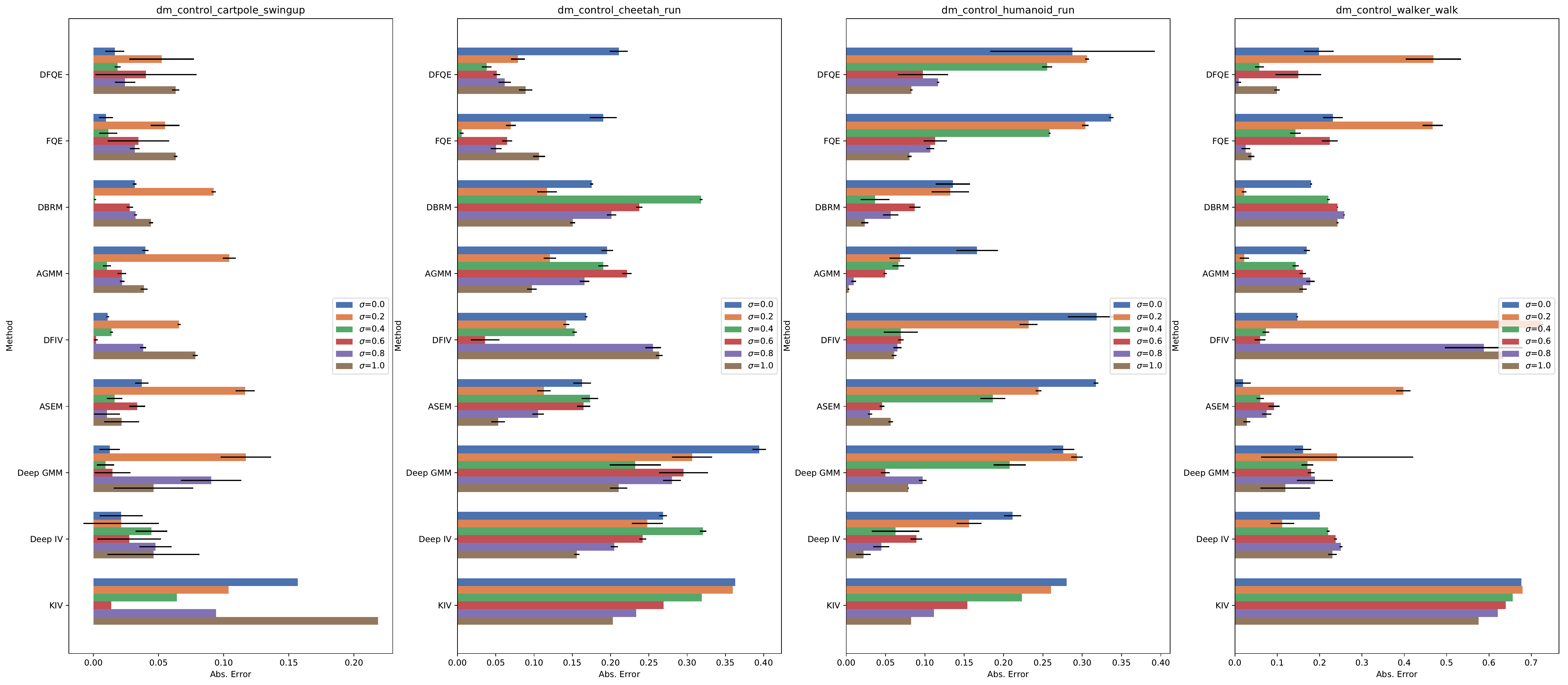}\\
    \caption{Absolute error of policy value estimation with the offline dataset}
    \label{fig:offline_policy_mae}
\end{figure}

\begin{figure}[tbhp!]
    \centering
    \includegraphics[width=0.6\textwidth]{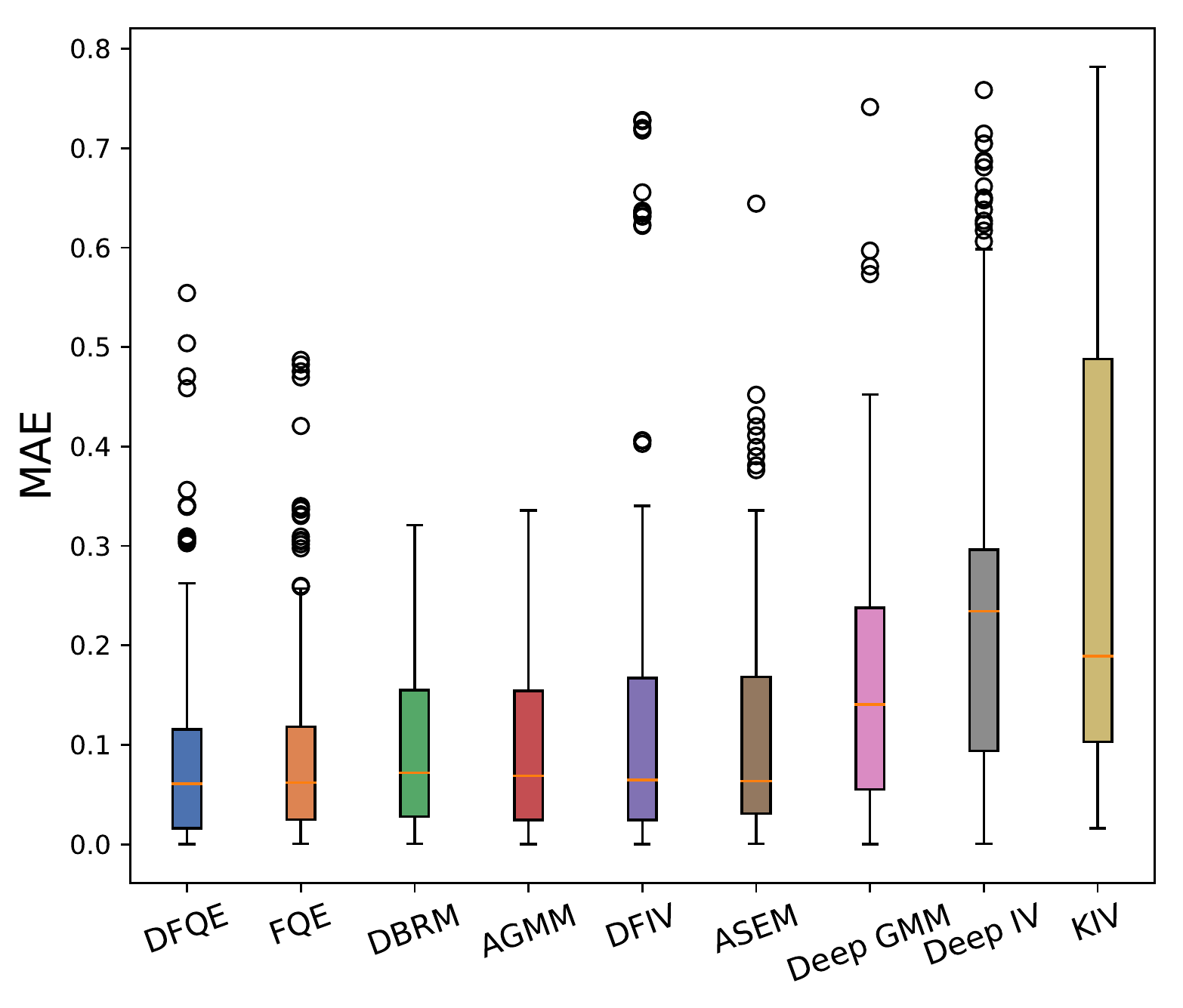}
    \caption{Distribution of the absolute error across all tasks with the offline dataset}
    \label{fig:offline_policy_mae_pool}
\end{figure}

We show the scatter plot of the estimated policy value versus ground truth in \Cref{fig:offline_policy_scatter}, the absolute error of the estimation per task and pooled together in \Cref{fig:offline_policy_mae} and \Cref{fig:offline_policy_mae_pool} respectively. 

Apparently, none of the algorithms considered here outperform all the other algorithms in all tasks. For example, DFQE and FQE perform relatively well in most tasks but fail on the Humanoid Run and Walker Walk control tasks. In contrast, AGMM estimates the policy value most accurately among all methods on the challenging Humanoid Run tasks but has higher error in BSuite catch, DM Control Cheetah Run and Walker Walk tasks. This observation is consistent with the results in \citep{xu2021learning} that benchmarks other OPE algorithms in deterministic environments, where they observe that ``no evaluated algorithm attains near-maximum performance under any metric''.

Nonetheless, by inspecting the estimation error in \Cref{fig:offline_policy_mae} and \Cref{fig:offline_policy_mae_pool} we notice that the relative performance of IV methods is roughly in agreement with the results observed in the near-policy dataset.
AGMM is the most robust IV method implemented here. ASEM and DFIV has a similar median error in all tasks but have a few large estimation errors in some tasks such as Walker Walk and BSuite Cartpole. Given the similarity between AGMM and ASEM, we conjecture that the relatively worse performance of ASEM is due to a bad choice of the hyper-parameters as it has more regularization hyper-parameters to tune.
KIV and Deep IV obtain considerably larger error than all the other methods, followed by Deep GMM, which has large variance in a few tasks such as Mountain Car and Walker Walk.

Surprisingly, DBRM that simply ignores the stochasticity of the dynamics displays the most  robust performance among all the algorithms. It converges quickly and almost always provides estimates of lowest variance among different runs compared to the other more complicated algorithms, at the price of a somewhat higher median error. 

Nonetheless, this observation is indeed understandable. The estimation error of the policy value comes from multiple sources including:
(1) the approximation error due to lack of model capacity (2) generalization error due to the limited dataset size (3) the off-policyness, or the divergence between the offline data distribution and the target policy distribution, which affects the effective size of the dataset (4) optimization error, affected by the optimization algorithm, and (5) the bias of ignoring the stochasticity of the dynamics.
As demonstrated in \Cref{Sec:toy_experiment}, the benefits of IV methods depend on multiple conditions.
Compared to the large near-policy dataset on BSuite, the pure offline dataset, including the continuous control tasks has higher modeling challenges, less data, and a larger divergence between the behavior and target distributions. All the other sources of error may dominate the bias present in DBRM, and due to the simplicity of DBRM as a single minimization problem, it is simpler to optimize than the loss of the other algorithms.


\section{Conclusion}\label{Sec:conclusion}
The regression problem one needs to solve when estimating a Q-function suffers from confounding as a result of the inputs and the output noise being correlated. If this confounding is ignored, one can obtain significantly biased Q-function estimates. We have shown here that fixing the target Q-network in DQN and FQE can be thought of as a way to overcome this confounding problem. As first suggested in \cite{xu2021learning}, another approach overcoming this problem consists of using IV regression techniques. 

By exploiting this observation and bringing together the literature on IV and RL, we have presented here various general nonlinear IV techniques developed recently in machine learning to estimate Q-functions parameterized by neural networks in the OPE context. This has allowed us not only to recover previously proposed OPE methods such as model-based techniques but also to obtain novel techniques. We have assessed the  performance of the resulting algorithms on a simple MDP model and two sets of benchmarking problems.

On the simple MDP problem, we find that the confounding effect can be very pronounced, and ignoring it as in  Deterministic Bellman Residual Minimization (DBRM) \citep{saleh2019} is problematic. In this example, an IV method like LSTD proves very useful. On more realistic examples from BSuite and DM Control, we have investigated scenarios where the evaluation policy is close to the behavioral one and where it is far from it. When doing OPE for a policy near to the one having generated data, we find that the confounding effect could be very pronounced and that techniques like DBRM are performing poorly. We also find that the best IV method - AGMM - displays performance on par with FQE and is only outperformed by distributional FQE. However, when evaluating a policy far from the one(s) having generated the observations, the combination of distribution shift and model mismatch has a non-negligible impact on performance. While we observe that AGMM also performed on par with FQE and DFQE, DBRM surprisingly also performs very well, and in general appears more stable (suffers from fewer outliers with poor performance) than FQE and DFQE. 

\noindent {\bf Acknowledgements} 

\noindent This work was supported by the Gatsby Charitable Foundation.

\bibliography{references}

\newpage

\appendix

\section{Variant of DFIV}
\label{Sec:dfiv_variant}

With the linear assumption of the $Q$ function, $Q_\theta(s,a) = \phi(s,a)^\top \theta$, we can follow the derivation of LSTD and write the structural function as
$$
f(s, a, s', a') = \left(\phi(s,a) - \gamma \phi(s',a')\right)^\top \theta\,.
$$
The original DFIV method exploits the decomposition of the structural function and regresses the expected deep features depending on $(s', a')$ only, $\expect{\phi(s',a')|s,a}$, in the first stage. We consider a variant where we regress the entire feature map directly $\expect{\phi(s,a) - \gamma \phi(s',a')|s, a} = V\psi(s, a)$, and obtain the following stage 1 regression,
\begin{align}
    \hat{V} = \argmin_{V} \mathcal{L}_{1}(V), \quad \mathcal{L}_{1}(V) = \expect[s,a,s',a']{\|\phi(s,a) - \gamma \phi(s',a') - V\psi(s,a) \|^2} + \lambda_1 \|V\|^2.
\end{align}
Stage 2 regression is then simplified as
\begin{align}
\hat{\theta} = \argmin_\theta \mathcal{L}_{2}(\theta), \quad \mathcal{L}_{2}(\theta) = \expect[r,s,a]{\|r - \theta^\top \hat{V}\psi(s,a) \|^2} + \lambda_2 \|\theta\|^2.
\label{eq:dfiv_variant_stage2_loss}
\end{align}
While both versions make use of instrumental variables, we find in the experiments that the second variant is more stable during training. A potential explanation is that the stage 1 target, $\phi(s,a) - \gamma \phi(s',a')$, has a smaller variance than $\phi(s,a)$ when the state-action pair $(s, a)$ changes smoothly in consecutive steps, and is easier to model by a least squares regression. Further investigation is yet to be done to verify this hypothesis.

\section{Hyper-parameter selection}
\label{Sec:hyper}

We choose the hyper-parameters of each IV algorithm using the recommended method in their original work. To be self-contained, we give a brief description for each method in this section. We split the offline dataset randomly into a training $\mathcal{D}_{\text{train}}$ and validation $\mathcal{D}_{\text{valid}}$ dataset with a ratio of 9:1, train each algorithm on $\mathcal{D}_{\text{train}}$ and choose the best hyper-parameters using $\mathcal{D}_{\text{valid}}$. The mini-batch size is 1024 unless explicitly specified.

Please note, however, that choosing the best hyper-parameters on the validation set does not guarantee a high accuracy in evaluating the target policy value, because the behavior distribution of states where the algorithm learns from is different from the evaluation distribution where the target policy induces and the learned OPE algorithm should be tested in. There is currently no commonly accepted way of choosing hyper-parameters in the offline reinforcement learning setting (see, e.g.\ \citet{paine2020hyperparameter}, for an attempt in choosing hyper-parameters for offline RL algorithms).

\subsection{Kernel Instrumental Variable}
We use random Fourier features to approximate the squared exponential kernel. The hyper-parameters of KIV are provided in \Cref{tab:hyper_kiv}. The validation metric for choosing the hyper-parameters is the stage-2 loss, \cref{eq:kiv_stage2_loss}, without regularization.

\begin{table}
    \centering
    \begin{tabular}{l|c}
        Hyper-parameter & Value range\\
        \hline
        Stage-1 regularization & $\{10^{-8}, 10^{-6}, 10^{-4}, 10^{-2}\}$ \\
        Stage-2 regularization & $\{10^{-8}, 10^{-6}, 10^{-4}, 10^{-2}\}$ \\
        Number of random features & $\{128, 256, 512, 1024\}$
    \end{tabular}
    \caption{Hyper-parameter of KIV. Values in braces are candidates to search over.}
    \label{tab:hyper_kiv}
\end{table}

\subsection{Deep Feature Instrumental Variables}

The hyper-parameters of DFIV include the $L_2$ regularization strength in the regression of both stages, $L_2$ regularization strength for the value and instrumental network parameters, the learning rate of the Adam optimizer, and the network architecture of the instrument network. The mini-batch size is fixed at 2048. Details are shown in \Cref{tab:hyper_dfiv}. The validation metric for choosing the hyper-parameters is the stage-2 loss in \Cref{eq:dfiv_variant_stage2_loss}, without regularization.

\begin{table}
    \centering
    \begin{tabular}{l|c}
        Hyper-parameter & Value range\\
        \hline
        Training steps & $10^5$ \\
        Stage-1 regularization & $\{10^{-8}, 10^{-6}, 10^{-4}, 10^{-2}\}$ \\
        Stage-2 regularization & $\{10^{-8}, 10^{-6}, 10^{-4}, 10^{-2}\}$ \\
        Value net regularization & $\{10^{-8}, 10^{-6}, 10^{-4}, 10^{-2}\}$ \\
        Instrument net regularization & $\{10^{-8}, 10^{-6}, 10^{-4}, 10^{-2}\}$ \\
        Value net learning rate & $\{10^{-5}, 3\times 10^{-5}, 10^{-4}, 3\times 10^{-4},10^{-3}\}$ \\
        Instrument net learning rate & $\{10^{-5}, 3\times 10^{-5}, 10^{-4}, 3\times 10^{-4},10^{-3}\}$ \\
        Instrument net hidden units (BSuite) & $\{(50, 50), (100, 100), (150, 150)\}$ \\
        Instrument net hidden units (DM Control) & $\{(512,512,256), (768,768,384), (1024,1024,512)\}$
    \end{tabular}
    \caption{Hyper-parameter of DFIV. Values in braces are candidates to search over.}
    \label{tab:hyper_dfiv}
\end{table}

\subsection{Deep IV}

The hyper-parameters of Deep IV include the hidden layer sizes and the number of mixing components in the treatment network, learning rate of treatment and value networks, and the number of Monte Carlo samples in estimating the integral in \Cref{eq:deep_iv}. The treatment network outputs both a mixture of Gaussian distribution to predict the next state $s'$ and a Bernoulli distribution to predict if the current state is a terminating state. Details are shown in \Cref{tab:hyper_deep_iv}. 

As suggested in \citet{hartford2017deep} we choose the hyper-parameters associated with training the treatment network according to the log-likelihood on the validation dataset, and those associated with training the value network according to the regression loss on the validation dataset.

\begin{table}
    \centering
    \begin{tabular}{p{2cm}|p{4cm}|p{5cm}}
        Network & Hyper-parameter & Value range\\
        \hline
        \multirow{5}{*}{\parbox{2cm}{Treatment net}} & Training steps & $10^5$ \\
        & Hidden units (BSuite) & $\{(32, 32), (64, 64), (128, 128)\}$ \\
        & Hidden units (DM Control) & $\{(128, 128, 128), (256, 256, 256)$, $(512,512,256), (768,768,384)\}$ \\
        & \# mixing components & $\{1, 3, 10\}$ \\
        & Learning rate & $\{10^{-5}, 3\times 10^{-5}, 10^{-4}, 3\times 10^{-4}, 10^{-3}, 3\times 10^{-3}\}$ \\
        \hline
        \multirow{3}{*}{\parbox{2cm}{Value net}} & Training steps & $10^5$ \\
        & \# samples & $\{1, 3, 10\}$ \\
        & Learning rate & $\{10^{-5}, 3\times 10^{-5}, 10^{-4}, 3\times 10^{-4}$, $10^{-3}, 3\times 10^{-3}\}$
    \end{tabular}
    \caption{Hyper-parameter of DFIV. Values in braces are candidates to search over.}
    \label{tab:hyper_deep_iv}
\end{table}

\subsection{Generalized Method of Moments}
\subsubsection{DeepGMM}

The hyper-parameters of DeepGMM include the hidden layer sizes of the adversarial network, learning rate of the value network $\eta_v$ and a learning rate multiplier for the adversarial network $\eta_a = \lambda \eta_v$, and the parameters $\beta_1$ and $\beta_2$ for the OAdam optimizer. Details are shown in \Cref{tab:hyper_deepgmm}. 

We follow the hyper-parameter selection method in \citet{bennett2019deep}. For every candidate of hyper-parameter setting and every checkpoint during the training $i$, we evaluate the value function $Q_{\theta_i}$ and adversarial function $g_{\tau_i}$ on a fixed set of validation data points. The metric for choosing the hyper-parameters is the objective in \Cref{eq:gmm_objective_q} except that it is evaluated on the validation set (both $\Psi_n$ and the expectation in $R_g$), the function $g$ is in the finite set $\{g_{\tau_i}\}$, and $Q_{\tilde{\theta}}$ in the regularization $R_g$ is averaged over all $Q_{\theta_i}$.
\begin{equation}
    \theta^* = \underset{\theta\in\{\theta_i\}}{\arg\min}\max_{\tau\in\{\tau_i\}} \Psi_n(Q_\theta, g_\tau) 
    - \frac{1}{4} \expect{g_\tau^2(s, a) (r - Q_{\tilde{\theta}}(s, a) + \gamma Q_{\tilde{\theta}}(s', a'))^2}\,,
\end{equation}
The hyper-parameters used to train $\theta^*$ is selected.

\begin{table}
    \centering
    \begin{tabular}{p{4cm}|c}
        Hyper-parameter & Value range\\
        \hline
        Training steps & $20^5$ \\
        Adversarial net hidden units (BSuite) & $\{(50, 50), (100, 100), (150, 150)\}$ \\
        Adversarial net hidden units (DM Control) & $\{(512,512,256), (768,768,384), (1024,1024,512)\}$ \\
        Value net learning rate & $\{10^{-5}, 3\times 10^{-5}, 10^{-4}, 3\times 10^{-4},10^{-3}\}$ \\
        Instrument net learning rate multiplier & $\{1, 5, 10, 50\}$\\
        OAdam $(\beta_1, \beta_2)$ & $\{(0, 0.01), (0.5, 0.9)\}$
    \end{tabular}
    \caption{Hyper-parameter of DeepGMM. Values in braces are candidates to search over.}
    \label{tab:hyper_deepgmm}
\end{table}

\subsubsection{Adversarial GMM and SEM}
The hyper-parameters of Adversarial GMM and SEM are similar to DeepGMM with additional hyper-parameters in the regularization terms. Details are shown in \Cref{tab:hyper_agmm} and \Cref{tab:hyper_asem}.

We choose the best hyper-parameter based on the early stopping method in the open-sourced implementation\footnote{\url{https://github.com/microsoft/AdversarialGMM/tree/main/mliv/neuralnet}} of AGMM. Particularly, we keep a set of candidate parameters $Q_{\theta_i}$ and $g_{\tau_i}$ and the validation set as in the previous section, and then find parameter $\theta_i$ with minimum moment violation
\begin{equation}
    \theta^* = \underset{\theta\in\{\theta_i\}}{\arg\min}\max_{\tau\in\{\tau_i\}} \Psi_n(Q_\theta, g_\tau)\,.
\end{equation}
In order to avoid the $\max$ operator to be dominated by a test function with a large magnitude, we normalize all the $g_{\tau_i}$ functions on the validation set, that is,
$$\tilde{g}(s, a) = \frac{1}{|\mathcal{D}_{\text{valid}}|}\expect[\mathcal{D}_{\text{valid}}]{g(s,a)}.$$

\begin{table}
    \centering
    \begin{tabular}{p{4cm}|c}
        Hyper-parameter & Value range\\
        \hline
        Training steps & $20^5$ \\
        Adversarial net hidden units (BSuite) & $\{(50, 50), (100, 100), (150, 150)\}$ \\
        Adversarial net hidden units (DM Control) & $\{(512,512,256), (768,768,384), (1024,1024,512)\}$ \\
        Value net learning rate & $\{10^{-5}, 3\times 10^{-5}, 10^{-4}, 3\times 10^{-4},10^{-3}\}$ \\
        Instrument net learning rate multiplier & $\{1, 5, 10, 50\}$ \\
        OAdam $(\beta_1, \beta_2)$ & $\{(0, 0.01), (0.5, 0.9)\}$ \\
        Value net parameter regularization $a$ & $\{10^{-10}, 10^{-8}, 10^{-6}, 10^{-4}, 10^{-2}\}$ \\
        Adversarial net parameter regularization $b$ & $\{10^{-10}, 10^{-8}, 10^{-6}, 10^{-4}, 10^{-2}\}$
    \end{tabular}
    \caption{Hyper-parameter of Adversarial GMM. Values in braces are candidates to search over.}
    \label{tab:hyper_agmm}
\end{table}

\begin{table}
    \centering
    \begin{tabular}{p{4cm}|c}
        Hyper-parameter & Value range\\
        \hline
        Training steps & $20^5$ \\
        Adversarial net hidden units (BSuite) & $\{(50, 50), (100, 100), (150, 150)\}$ \\
        Adversarial net hidden units (DM Control) & $\{(512,512,256), (768,768,384), (1024,1024,512)\}$ \\
        Value net learning rate & $\{10^{-5}, 3\times 10^{-5}, 10^{-4}, 3\times 10^{-4},10^{-3}\}$ \\
        Instrument net learning rate multiplier & $\{1, 5, 10, 50\}$ \\
        OAdam $(\beta_1, \beta_2)$ & $\{(0, 0.01), (0.5, 0.9)\}$ \\
        Value net parameter regularization $a$ & $\{10^{-8}, 10^{-6}, 10^{-4}, 10^{-2}\}$ \\
        Adversarial net parameter regularization $b$ & $\{10^{-8}, 10^{-6}, 10^{-4}, 10^{-2}\}$ \\
        Value net parameter regularization $\alpha$ & $\{10^{-8}, 10^{-6}, 10^{-4}, 10^{-2}\}$
    \end{tabular}
    \caption{Hyper-parameter of Adversarial SEM. Values in braces are candidates to search over.}
    \label{tab:hyper_asem}
\end{table}

\end{document}